\documentclass[10pt,twocolumn,letterpaper]{article}

\usepackage{iccv}
\usepackage{times}
\usepackage{epsfig}
\usepackage{graphicx}
\usepackage{amsmath}
\usepackage{amssymb}

\usepackage{epstopdf}
\usepackage{enumitem}

\usepackage{amsmath,amsfonts}
\usepackage{amssymb,amsopn}
\usepackage{bm} 

\usepackage{amssymb}
\usepackage{amsmath,amsfonts}
\usepackage{amsthm,amsopn}

\usepackage{bm} 

\newcommand{\mt}[1]{{\small\textsc{#1}}}
\newcommand{\mtc}[1]{{\footnotesize\textsc{#1}}}
\newcommand{\mtt}[1]{{\scriptsize\textsc{#1}}}

\newcommand{\vct}[1]{\boldsymbol{#1}} 
\newcommand{\mat}[1]{\boldsymbol{#1}} 
\newcommand{\cst}[1]{\mathsf{#1}}  

\newcommand{\field}[1]{\mathbb{#1}}
\newcommand{\R}{\field{R}} 

\newcommand{\twonorm}[1]{\left\|#1\right\|_2^2}

\newcommand{\ProbOpr}[1]{\mathbb{#1}}

\newcommand{\expect}[2]{%
\ifthenelse{\equal{#2}{}}{\ProbOpr{E}_{#1}}
{\ifthenelse{\equal{#1}{}}{\ProbOpr{E}\left[#2\right]}{\ProbOpr{E}_{#1}\left[#2\right]}}} 
\newcommand{\var}[2]{%
\ifthenelse{\equal{#2}{}}{\ProbOpr{VAR}_{#1}}
{\ifthenelse{\equal{#1}{}}{\ProbOpr{VAR}\left[#2\right]}{\ProbOpr{VAR}_{#1}\left[#2\right]}}} 

\newcommand{\vtheta}{\vct{\theta}}

\newcommand{\vx}{{\vct{x}}}

\newcommand{\va}{\vct{a}}
\newcommand{\vb}{\vct{b}}

\newcommand{\vv}{\vct{v}}
\newcommand{\vw}{\vct{w}}

\newcommand{\vpsi}{\vct{\psi}}

\newcommand{\mW}{\mat{W}}

\newcommand{\mM}{\mat{M}}

\newcommand{\mD}{\mat{D}}

\newcommand{\cN}{\cst{N}}

\newcommand{\cD}{\cst{D}}

\newcommand{\cR}{\cst{R}}
\newcommand{\cU}{\cst{U}}
\newcommand{\cS}{\cst{S}}

\newcommand{\eat}[1]{}

\usepackage[pagebackref=true,breaklinks=true,colorlinks,bookmarks=false]{hyperref}

\iccvfinalcopy 

\def\iccvPaperID{1472} 

\ificcvfinal\fi
\begin{document}

\title{Predicting Visual Exemplars of Unseen Classes for Zero-Shot Learning}

\author{
Soravit Changpinyo\\
U. of Southern California\\
Los Angeles, CA\\
{\tt\small schangpi@usc.edu}
\and
Wei-Lun Chao\\
U. of Southern California\\
Los Angeles, CA\\
{\tt\small weilun@usc.edu}
\and
Fei Sha\\
U. of Southern California\\
Los Angeles, CA\\
{\tt\small feisha@usc.edu}
}

\maketitle


\begin{abstract}
Leveraging class semantic descriptions and examples of known objects, zero-shot learning makes it possible to train a recognition model for an object class whose examples are not available. In this paper, we propose a novel zero-shot learning model that takes advantage of clustering structures in the semantic embedding space.
The key idea is to impose the structural constraint that semantic representations must be predictive of the locations of their corresponding visual \emph{exemplars}. To this end, this reduces to training multiple kernel-based regressors from semantic representation-exemplar pairs from labeled data of the seen object categories. Despite its simplicity, our approach significantly outperforms existing zero-shot learning methods on standard benchmark datasets, including the ImageNet dataset with more than 20,000 unseen categories.
\end{abstract}


\section{Introduction}
\label{sIntro}

\begin{figure*}[ht]
\centering
\includegraphics[width=0.7\textwidth]{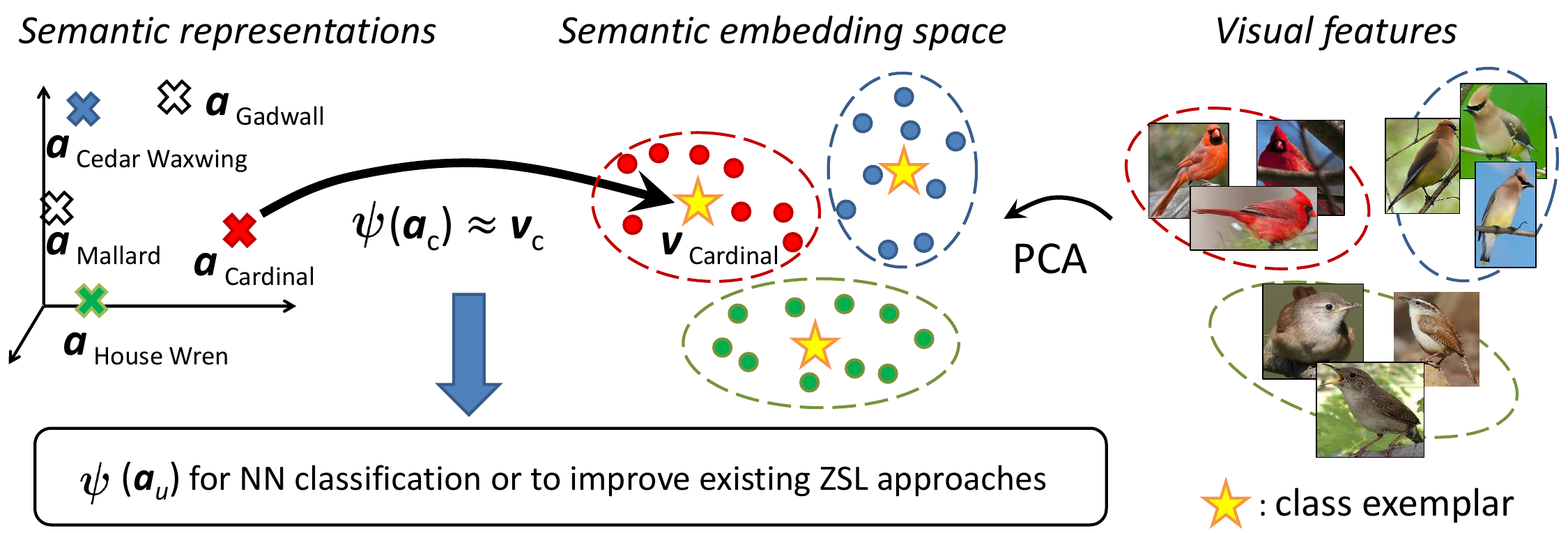}
\caption{\small Given the semantic information and visual features of the seen classes, our method learns a \textbf{kernel-based regressor $\vpsi(\cdot)$} such that the semantic representation $\va_c$ of class $c$ can \textbf{predict well} its class exemplar (center) $\vv_c$ that characterizes the clustering structure. The learned $\vpsi(\cdot)$ can be used to predict the visual feature vectors of the unseen classes for nearest-neighbor (NN) classification, or to improve the semantic representations for existing ZSL approaches.} 
\label{fConcept} 
\vspace{-16pt}
\end{figure*}

A series of major progresses in visual object recognition can largely be attributed to learning large-scale and complex models with a huge number of labeled training images. There are many application scenarios, however, where collecting and labeling training instances can be laboriously difficult and costly. For example, when the objects of interest are rare (e.g., only about a hundred of northern hairy-nosed wombats alive in the wild) or newly defined (e.g., images of futuristic products such as Tesla's Model S), not only the amount of the labeled training images but also the statistical variation among them is limited. These restrictions do not lead to robust systems for recognizing such objects. More importantly, the number of such objects could be significantly greater than the number of common objects. In other words, the frequencies of observing objects follow a long-tailed distribution~\cite{SalakhutdinovTT11,ZhuAR14}.
 
Zero-shot learning (ZSL) has since emerged as a promising paradigm to remedy the above difficulties. Unlike supervised learning, ZSL distinguishes between two types of classes: \emph{seen} and \emph{unseen}, where labeled examples are available for the seen classes only. Crucially, zero-shot learners have access to a shared semantic space that embeds all categories. This semantic space enables transferring and adapting classifiers trained on the seen classes to the unseen ones. Multiple types of semantic information have been exploited in the literature: visual attributes \cite{FarhadiEHF09,LampertNH09}, word vector representations of class names \cite{FromeCSBDRM13,SocherGMN13,NorouziMBSSFCD14}, textual descriptions \cite{ElhoseinySE13,LeiSFS15,ReedKLS16}, hierarchical ontology of classes (such as WordNet \cite{Miller95}) \cite{AkataRWLS15,Lu16,XianASNHS16}, and human gazes \cite{KaressliABS17}.

Many ZSL methods take a two-stage approach: (i) predicting the embedding of the image in the semantic space;  (ii) inferring the class labels by comparing the embedding to the unseen classes' semantic representations~\cite{FarhadiEHF09,LampertNH09,PalatucciPHM09,SocherGMN13,YuCFSC13,JayaramanG14,NorouziMBSSFCD14,Lu16}. Recent ZSL methods take a unified approach by jointly learning the functions to predict the semantic embeddings as well as to measure similarity in the embedding space~\cite{AkataPHS13,AkataRWLS15,FromeCSBDRM13,Bernardino15,ZhangS15,ZhangS16,ChangpinyoCGS16}. We refer the readers to the descriptions and evaluation on these representative methods in \cite{XianAS17}.

Despite these attempts, zero-shot learning is proved to be extremely difficult. For example, the best reported accuracy on the full ImageNet with 21K categories is only 1.5\% \cite{ChangpinyoCGS16}, where the state-of-the-art performance with supervised learning reaches 29.8\%~\cite{ChilimbiSAK14}\footnote{Comparison between the two numbers is not entirely fair due to different training/test splits. Nevertheless, it gives us a rough idea on how huge the gap is. This observation has also been shown on small datasets \cite{ChaoCGS16}.}. 

There are at least two critical reasons for this.
First, class semantic representations are vital for knowledge transfer from the seen classes to unseen ones, but these representations are hard to get right. Visual attributes are human-understandable so they correspond well with our object class definition. However, they are not always discriminative \cite{ParikhG11a,YuCFSC13}, not necessarily machine detectable \cite{DuanPCG12,JayaramanG14}, often correlated among themselves (``brown" and ``wooden"') \cite{JayaramanSG14}, and possibly not category-independent (``fluffy'' animals and ``fluffy" towels) \cite{ChenG14}. Word vectors of class names have shown to be inferior to attributes \cite{AkataRWLS15,ChangpinyoCGS16}. 
Derived from texts, they have little knowledge about or are barely aligned with visual information.
 
The other reason is that the lack of data for the unseen classes presents a unique challenge for model selection.
The crux of ZSL involves learning a compatibility function between the visual feature of an image and the semantic representation of each class. But, how are we going to parameterize this function?
Complex functions are flexible but at risk of overfitting to the seen classes and transferring poorly to the unseen ones. Simple ones, on the other hand, will result in poorly performing classifiers on the seen classes and will unlikely perform well either on the unseen ones. 
For these reasons, the success of ZSL methods hinges critically on the insight of the underlying mechanism for transfer and how well that insight is in accordance with data.

One particular fruitful (and often implicitly stated) insight is the existence of clustering structures in the semantic embedding space. 
That is, images of the same class, after embedded into the semantic space, will cluster around the semantic embedding of that class.
For example, ConSE \cite{NorouziMBSSFCD14} aligns a convex composition of the classifier probabilistic outputs to the semantic representations. 
A recent method of synthesized classifiers (SynC) \cite{ChangpinyoCGS16} models two aligned manifolds of clusters, one corresponding to the semantic embeddings of all objects and the other corresponding to the ``centers''\footnote{The centers are defined as the normals of the hyperplanes separating different classes.} in the visual feature space, where the pairwise distances between entities in each space are used to constrain the shapes of both manifolds. These lines of insights have since yielded excellent performance on ZSL.

In this paper, we propose a simple yet very effective ZSL algorithm that assumes and leverages more \emph{structural relations} on the clusters.
The main idea is to exploit the intuition that the semantic representation can \textbf{predict well} the location of the cluster characterizing all visual feature vectors from the corresponding class (c.f. Sect. \ref{sec_vis_exem}).

More specifically, the main computation step of our approach is reduced to learning (from the seen classes) a predictive function from semantic representations to their corresponding centers (i.e., \emph{exemplars}) of visual feature vectors.
This function is used to predict the locations of visual exemplars of the unseen classes that are then used to construct nearest-neighbor style classifiers, or to improve the semantic information demanded by existing ZSL approaches. Fig.~\ref{fConcept} shows the conceptual diagram of our approach.

Our proposed method tackles the two challenges for ZSL simultaneously. 
First, unlike most of the existing ZSL methods, we acknowledge that semantic representations may not necessarily contain visually discriminating properties of objects classes. As a result, we demand that the \emph{predictive} constraint be imposed explicitly. In our case, we assume that the cluster centers of visual feature vectors are our \emph{target} semantic representations. 
Second, we leverage structural relations on the clusters to further regularize the model, strengthening the usefulness of the clustering structure assumption for model selection.

We validate the effectiveness of our proposed approach on four benchmark datasets for ZSL, including the full ImageNet dataset with more than 20,000 unseen classes. Despite its simplicity, our approach outperforms other existing ZSL approaches in most cases, demonstrating the potential of exploiting the structural relatedness between visual features and semantic information.
Additionally, we complement our empirical studies with extensions from zero-shot to few-shot learning, as well as analysis of our approach. 

The rest of the paper is organized as follows. We describe our proposed approach in Sect.~\ref{EXEM}. We demonstrate the superior performance of our method in Sect.~\ref{sExp}. We discuss relevant work in Sect.~\ref{sRelated} and finally conclude in Sect.~\ref{sDiscuss}.

\section{Approach}
\label{EXEM}

We describe our methods for addressing zero-shot learning, where the task is to classify images from the unseen classes into the label space of the unseen classes.
Our approach is based on the structural constraint that takes advantage of the clustering structure assumption in the semantic embedding space.
The constraint forces the semantic representations to be predictive of their visual exemplars (i.e., cluster centers).
In this section, we describe how we achieve this goal. 
First, we describe how we learn a function to predict the visual exemplars from the semantic representations.
Second, given a novel semantic representation, we describe how we apply this function to perform zero-shot learning.

\paragraph{Notations} We follow the notation system introduced in \cite{ChangpinyoCGS16} to facilitate comparison. We denote by $\mathcal{D}= \{(\vx_n\in \R^{\cD},y_n)\}_{n=1}^\cN$ the training data with the labels from the label space of \emph{seen} classes $\mathcal{S} = \{1,2,\cdots,\cS\}$. we denote by $\mathcal{U} = \{\cS+1,\cdots,\cS+\cU\}$ the label space of \emph{unseen} classes. For each class $c \in \mathcal{S} \cup \mathcal{U}$, let $\va_c$ be its semantic representation.

\subsection{Learning to predict the visual exemplars from the semantic representations}
\label{sSVR}
For each class $c$, we would like to find a transformation function $\vpsi(\cdot)$ such that $\vpsi(\va_c) \approx \vv_c$, where $\vv_c \in \R^\cst{d}$ is the visual exemplar for the class. 
In this paper, we create the visual exemplar of a class by averaging the PCA projections of data belonging to that class. That is, we consider $\vv_c = \frac{1}{|I_c|}\sum_{n \in I_c} \mM \vx_n$, where $I_c = \{i: y_i = c\}$ and $\mM \in \R^{\cst{d} \times \cD} $ is the PCA projection matrix computed over training data of the seen classes. We note that $\mM$ is fixed for all data points (i.e., not class-specific) and is used in~Eq. (\ref{ePredict}).

Given training visual exemplars and semantic representations, we learn $\cst{d}$ support vector regressors (SVR) with the RBF kernel --- each of them predicts each dimension of visual exemplars from their corresponding semantic representations. Specifically, for each dimension $d = 1, \ldots, \cst{d}$, we use the $\nu$-SVR formulation \cite{ScholkopfSWB00}. Details are in the supplementary material.
 
Note that the PCA step is introduced for both the computational and statistical benefits. In addition to reducing dimensionality for faster computation, PCA decorrelates the dimensions of visual features such that we can predict these dimensions independently rather than jointly.

See Sect.~\ref{sec_analysis_svr_pca} for analysis on applying SVR and PCA.

\subsection{Zero-shot learning based on the predicted visual exemplars}
Now that we learn the transformation function $\vpsi(\cdot)$, how do we use it to perform zero-shot classification?
We first apply $\vpsi(\cdot)$ to all semantic representations $\va_u$ of the unseen classes.
We consider two main approaches that depend on how we interpret these predicted exemplars $\vpsi(\va_u)$.
\subsubsection{Predicted exemplars as training data} 
\label{sProto}
An obvious approach is to use $\vpsi(\va_u)$ as data directly. Since there is only one data point per class, a natural choice is to use a nearest neighbor classifier. Then, the classifier outputs the label of the closest exemplar for each novel data point $\vx$ that we would like to classify:
\begin{align}
\hat{y} = \arg\min_{u} \quad dis_{NN}(\mM\vx, \vpsi(\va_u)), 
\label{ePredict}
\end{align}
where we adopt the (standardized) Euclidean distance as $dis_{NN}$ in the experiments.

\subsubsection{Predicted exemplars as the ideal semantic representations}
\label{sIdeal}
The other approach is to use $\vpsi(\va_u)$ as the \emph{ideal} semantic representations (``ideal" in the sense that they have knowledge about visual features) and plug them into any existing zero-shot learning framework. We provide two examples.

In the method of convex combination of semantic embeddings (ConSE) \cite{NorouziMBSSFCD14}, their original semantic embeddings are replaced with the corresponding predicted exemplars, while the combining coefficients remain the same.  
In the method of synthesized classifiers (SynC) \cite{ChangpinyoCGS16}, the predicted exemplars are used to define the similarity values between the unseen classes and the bases, which in turn are used to compute the combination weights for constructing classifiers. In particular, their similarity measure is of the form $\frac{\exp\{-dis(\va_c,\vb_r)\}}{\sum_{r=1}^{\cR}\exp\{-dis(\va_c,\vb_r)\}}$, where $dis$ is the (scaled) Euclidean distance and $\vb_r$'s are the semantic representations of the base classes. In this case, we simply need to change this similarity measure to $\frac{\exp\{-dis(\vpsi(\va_c),\vpsi(\vb_r))\}}{\sum_{r=1}^{\cR}\exp\{-dis(\vpsi(\va_c),\vpsi(\vb_r))\}}$.

We note that, recently, Chao et al. \cite{ChaoCGS16} empirically show that existing semantic representations for ZSL are far from the optimal.
Our approach can thus be considered as a way to improve semantic representations for zero-shot learning.

\subsection{Comparison to related approaches}
\label{sEXEMcompare}
One appealing property of our approach is its scalability: we learn and predict at the exemplar (class) level so the runtime and memory footprint of our approach depend only on the number of seen classes rather the number of training data points. This is much more efficient than other ZSL algorithms that learn at the level of each individual training instance~\cite{FarhadiEHF09,LampertNH09,PalatucciPHM09,AkataPHS13,YuCFSC13,FromeCSBDRM13,SocherGMN13,NorouziMBSSFCD14,JayaramanG14,MensinkGS14,AkataRWLS15,Bernardino15,ZhangS15,ZhangS16,Lu16,ChangpinyoCGS16}.

Several methods propose to learn visual exemplars\footnote{\emph{Exemplars} are used loosely here and do not necessarily mean class-specific feature averages.} by preserving structures obtained in the semantic space \cite{ChangpinyoCGS16,WangC16,LongLSSDH17}. However, our approach \emph{predicts} them with a regressor such that they may or may not strictly follow the structure in the semantic space, and thus they are more flexible and could even better reflect similarities between classes in the visual feature space.

Similar in spirit to our work, \cite{MensinkVPC13} proposes using nearest class mean classifiers for ZSL. The Mahalanobis metric learning in this work could be thought of as learning a linear transformation of semantic representations (their ``zero-shot prior" means, which are in the visual feature space). Our approach learns a highly non-linear transformation. Moreover, our \mt{EXEM (1NNs)} (cf. Sect.~\ref{sExpSetup}) learns a (simpler, i.e., diagonal) metric over the learned exemplars. Finally, the main focus of \cite{MensinkVPC13} is on \emph{incremental}, not zero-shot, learning settings (see also \cite{RistinGGB16,RebuffiKL17}).

\cite{ZhangXG16} proposes to use a deep feature space as the semantic embedding space for ZSL. Though similar to ours, they do not compute average of visual features (exemplars) but train neural networks to predict \emph{all} visual features from their semantic representations. Their model learning takes significantly longer time than ours. Neural networks are more prone to overfitting and give inferior results (cf. Sect.~\ref{sec_analysis_svr_pca}). Additionally, we provide empirical studies on much larger-scale datasets for both zero-shot and few-shot learning, and analyze the effect of PCA. 

\section{Experiments}
\label{sExp}


We evaluate our methods and compare to existing state-of-the-art models on four benchmark datasets with diverse domains and scales. 
Despite variations in datasets, evaluation protocols, and implementation details, we aim to provide a comprehensive and fair comparison to existing methods by following the evaluation protocols in \cite{ChangpinyoCGS16}. Note that \cite{ChangpinyoCGS16} reports results of many other existing ZSL methods based on their settings.
Details on these settings are described below and in the supplementary material.

\subsection{Setup}
\label{sExpSetup}

\begin{table}
\centering
{\footnotesize
\caption{\small Key characteristics of the datasets}
\vskip .5em
\label{tDatasets}
 \begin{tabular}{c|c|c|c}
Dataset & \# of seen classes & \# of unseen classes & \# of images \\ \hline
\textbf{AwA}$^\dagger$ & 40  & 10 & 30,475\\ \hline
\textbf{CUB}$^\ddagger$ & 150 & 50 & 11,788\\ \hline
\textbf{SUN}$^\ddagger$ & 645/646  & 72/71 &  14,340\\ \hline
\textbf{ImageNet}$^\S$ & 1,000 & 20,842 & 14,197,122\\ \hline
\end{tabular}
\begin{flushleft}
$^\dagger$: on the prescribed split in~\cite{LampertNH14}.\\
$^\ddagger$: on 4 (or 10, respectively) random splits~\cite{ChangpinyoCGS16}, reporting average.\\
$^\S$: Seen and unseen classes from ImageNet ILSVRC 2012 1K~\cite{ILSVRC15} and Fall 2011 release~\cite{deng2009imagenet,FromeCSBDRM13,NorouziMBSSFCD14}.
\end{flushleft}
}
\vskip -2.5em
\end{table}

\paragraph{Datasets} 
We use four benchmark datasets for zero-shot learning in our experiments: \textbf{Animals with Attributes (AwA)}~\cite{LampertNH14},  \textbf{CUB-200-2011 Birds (CUB)}~\cite{WahCUB_200_2011}, \textbf{SUN Attribute (SUN)}~\cite{PattersonH14}, and \textbf{ImageNet} (with full 21,841 classes)~\cite{ILSVRC15}. Table~\ref{tDatasets} summarizes their key characteristics. The supplementary material provides more details.\\[0.25em]   
\noindent
\textbf{Semantic representations} We use the publicly available 85, 312, and 102 dimensional continuous-valued attributes for \textbf{AwA}, \textbf{CUB}, and \textbf{SUN}, respectively. For \textbf{ImageNet}, there are two types of semantic representations of the class names.
First, we use the 500 dimensional word vectors \cite{ChangpinyoCGS16} obtained from training a skip-gram model~\cite{MikolovCCD13} on Wikipedia. We remove the class names without word vectors, making the number of unseen classes to be 20,345 (out of 20,842). 
Second, we derive 21,632 dimensional semantic vectors of the class names using multidimensional scaling (MDS) on the WordNet hierarchy, as in \cite{Lu16}. 
We normalize the class semantic representations to have unit $\ell_2$ norms.\\[0.25em]   
\noindent
\textbf{Visual features} We use GoogLeNet features (1,024 dimensions)~\cite{SzegedyLJSRAEVR14} provided by \cite{ChangpinyoCGS16} due to their superior performance \cite{AkataRWLS15,ChangpinyoCGS16} and prevalence in existing literature on ZSL.\\[0.25em]   
\noindent
\textbf{Evaluation protocols} For \textbf{AwA}, \textbf{CUB}, and \textbf{SUN}, we use the multi-way classification accuracy (averaged over classes) as the evalution metric. On \textbf{ImageNet}, we describe below additional metrics and protocols introduced in \cite{FromeCSBDRM13} and followed by \cite{ChangpinyoCGS16,Lu16,NorouziMBSSFCD14}.

First, two evaluation metrics are employed: Flat hit@K (F@K) and Hierarchical precision@K (HP@K). F@K is defined as the percentage of test images for which the model returns the true label in its top K predictions. Note that, F@1 is the multi-way classification accuracy (averaged over samples). 
HP@K is defined as the percentage of overlapping (i.e., precision) between the model's top K predictions and the ground-truth list. For each class, the ground-truth list of its K closest categories is generated based on the ImageNet hierarchy~\cite{deng2009imagenet}. See the Appendix of \cite{FromeCSBDRM13,ChangpinyoCGS16} for details. Essentially, this metric allows for some errors as long as the predicted labels are semantically similar to the true one. 

Second, we evaluate ZSL methods on three subsets of the test data of increasing difficulty: \emph{2-hop}, \emph{3-hop}, and \emph{All}.
\emph{2-hop} contains 1,509 (out of 1,549) unseen classes that are within 2 tree hops of the 1K seen classes according to the ImageNet hierarchy.
\emph{3-hop} contains 7,678 (out of 7,860) unseen classes that are within 3 tree hops of seen classes. Finally, \emph{All} contains all 20,345 (out of 20,842) unseen classes in the ImageNet 2011 21K dataset that are not in the ILSVRC 2012 1K dataset.

Note that word vector embeddings are missing for certain class names with rare words.
For the MDS-WordNet features, we provide results for \emph{All} only for comparison to \cite{Lu16}.
In this case, the number of unseen classes is 20,842.\\[0.25em]   
\noindent
\textbf{Baselines}
We compare our approach with several state-of-the-art and recent competitive ZSL methods summarized in Table~\ref{tbMain}.
Our main focus will be on \mt{SynC} \cite{ChangpinyoCGS16}, which has recently been shown to have superior performance against competitors under the same setting, especially on large-scale datasets \cite{XianAS17}.
Note that \mt{SynC} has two versions: one-versus-other loss formulation \mt{SynC}$^\textrm{o-v-o}$ and the Crammer-Singer formulation~\cite{CrammerS02} \mt{SynC}$^\textrm{struct}$.
On small datasets, we also report results from recent competitive baselines \mt{LatEm} \cite{XianASNHS16} and \mt{BiDiLEL} \cite{WangC16}. For additional details regarding other (weaker) baselines, see the supplementary material.
Finally, we compare our approach to all ZSL methods that provide results on \textbf{ImageNet}. When using word vectors of the class names as semantic representations, we compare our method to \mt{ConSE} \cite{NorouziMBSSFCD14} and \mt{SynC} \cite{ChangpinyoCGS16}. When using MDS-WordNet features as semantic representations, we compare our method to \mt{SynC} \cite{ChangpinyoCGS16} and \mt{CCA} \cite{Lu16}.\\[0.25em]   
\noindent\textbf{Variants of our ZSL models given predicted exemplars}
The main step of our method is to predict visual exemplars that are well-informed about visual features.
How we proceed to perform zero-shot classification (i.e., classifying test data into the label space of unseen classes) based on such exemplars is entirely up to us.
In this paper, we consider the following zero-shot classification procedures that take advantage of the predicted exemplars:  
\vspace{-5pt}
\begin{itemize}[noitemsep]
\item \mt{EXEM (\emph{ZSL method}):} ZSL method with predicted exemplars as semantic representations, where \emph{ZSL method} $=$ \mt{ConSE} \cite{NorouziMBSSFCD14}, \mt{LatEm} \cite{XianASNHS16}, and \mt{SynC} \cite{ChangpinyoCGS16}.
\item \mt{EXEM (1NN):} 1-nearest neighbor classifier with the Euclidean distance to the exemplars.
\item \mt{EXEM (1NNs):} 1-nearest neighbor classifier with the \emph{standardized} Euclidean distance to the exemplars, where the standard deviation is obtained by averaging the intra-class standard deviations of all seen classes.
\end{itemize}

\mt{EXEM (\emph{ZSL method})} regards the predicted exemplars as the ideal semantic representations (Sect.~\ref{sIdeal}).
On the other hand, \mt{EXEM (1NN)} treats predicted exemplars as data prototypes (Sect.~\ref{sProto}).
The standardized Euclidean distance in \mt{EXEM (1NNs)} is introduced as a way to scale the variance of different dimensions of visual features. In other words, it helps reduce the effect of \emph{collapsing} data that is caused by our usage of the average of each class' data as cluster centers.\\[0.25em]   
\noindent
\textbf{Hyper-parameter tuning} We simulate zero-shot scenarios to perform 5-fold cross-validation during training. Details are in the supplementary material.

\subsection{Predicted visual exemplars}
\label{sec_vis_exem}
\begin{table}
\centering
{\footnotesize
\caption{\small We compute the Euclidean distance matrix between the \emph{unseen} classes based on semantic representations ($\mD_{\va_u}$), predicted exemplars ($\mD_{\vpsi(\va_u)}$), and real exemplars ($\mD_{\vv_u}$). Our method leads to $\mD_{\vpsi(\va_u)}$ that is better correlated with $\mD_{\vv_u}$ than $\mD_{\va_u}$ is. See text for more details.}
\vskip .5em
\label{tCorrDis}
\begin{tabular}{c|c|c}
Dataset & \multicolumn{2}{|c}{Correlation to $\mD_{\vv_u}$}  \\ \cline{2-3}
name & Semantic distances & Predicted exemplar distances\\
& $\mD_{\va_u}$ & $\mD_{\vpsi(\va_u)}$ \\ \hline
\textbf{AwA} & 0.862 & \textbf{0.897} \\ \hline
\textbf{CUB} & 0.777 $\pm$ 0.021 & \textbf{0.904} $\pm$ 0.026 \\ \hline
\textbf{SUN} & 0.784 $\pm$ 0.022 & \textbf{0.893} $\pm$ 0.019 \\ \hline
\end{tabular}
}
\vskip -1em
\end{table}

\begin{figure*}[ht]
\centering
\includegraphics[width=0.23\textwidth]{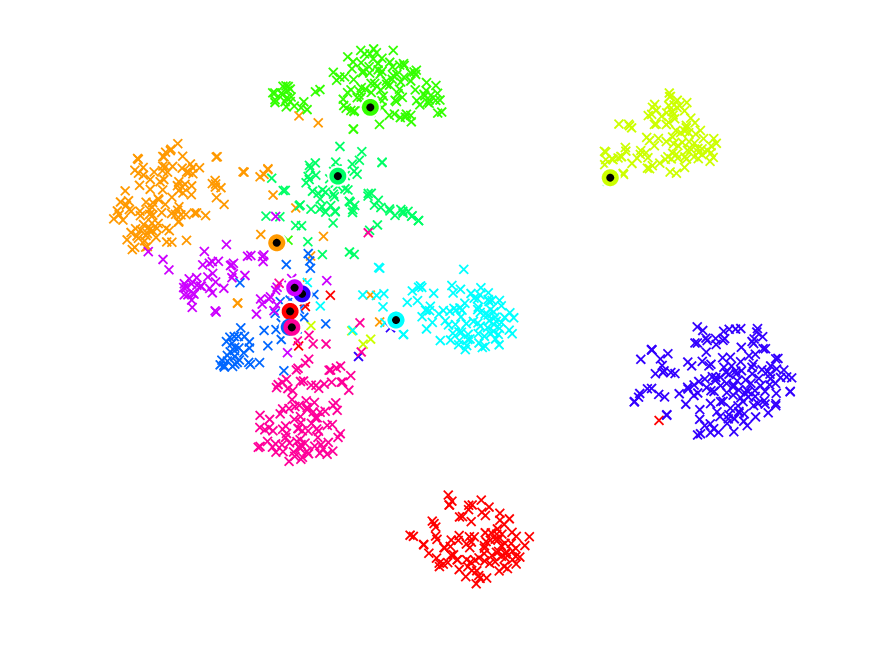}
\includegraphics[width=0.23\textwidth]{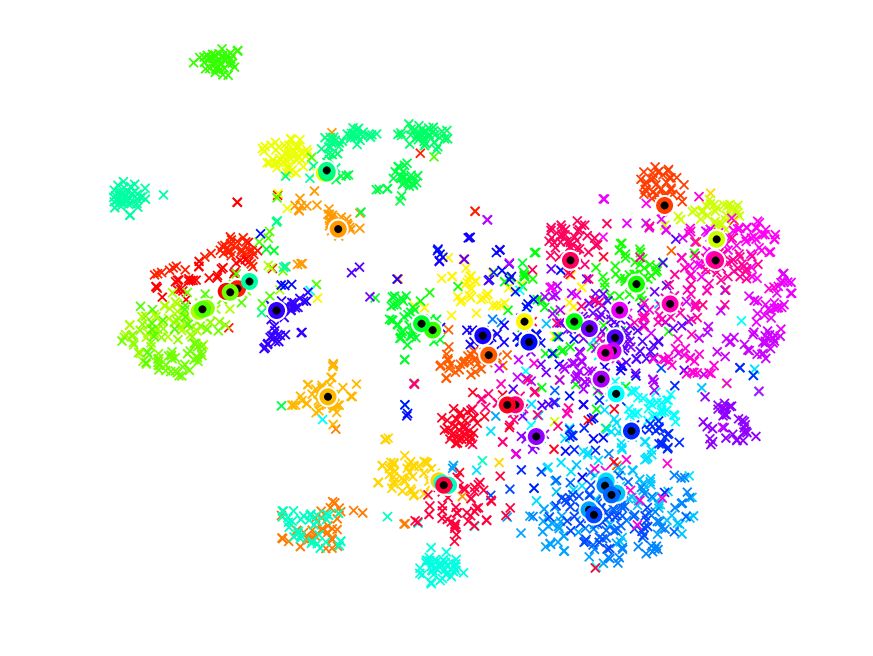}
\includegraphics[width=0.23\textwidth]{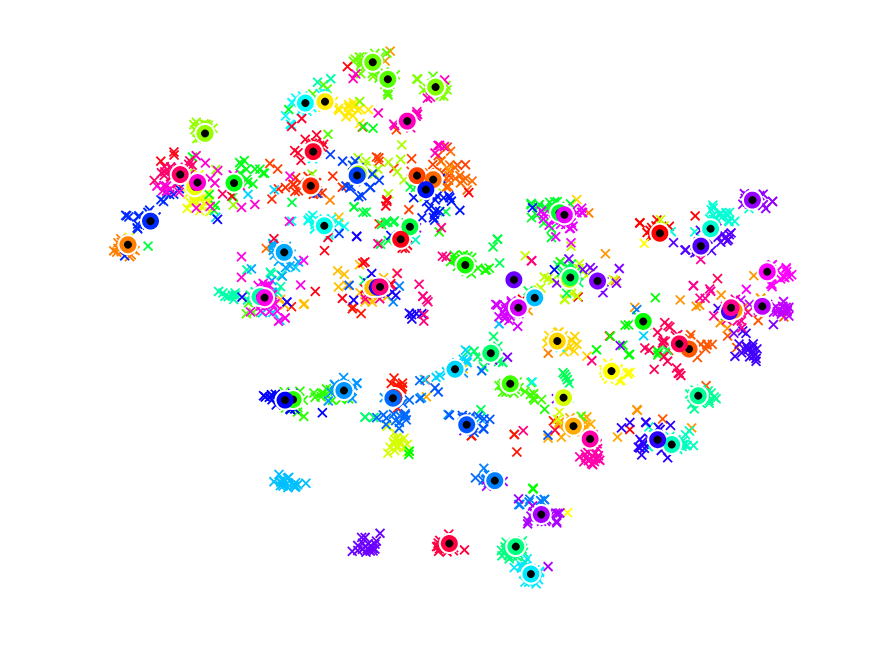}
\includegraphics[width=0.26\textwidth]{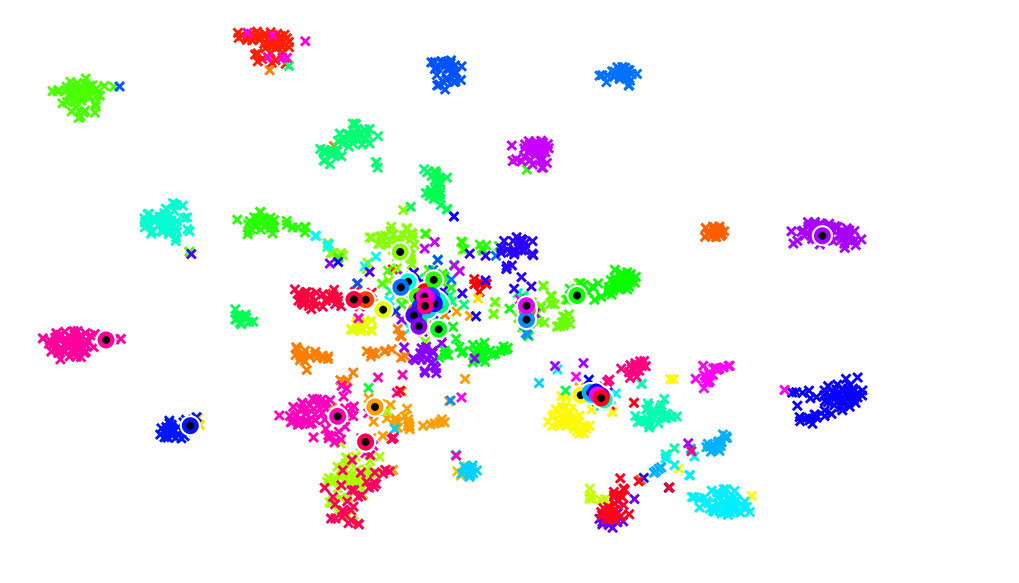}
\vspace{-7pt}
\caption{\small t-SNE \cite{tSNE} visualization of randomly selected real images (crosses) and predicted visual exemplars (circles) for the \emph{unseen} classes on (from left to right) \textbf{AwA}, \textbf{CUB}, \textbf{SUN}, and \textbf{ImageNet}. Different colors of symbols denote different unseen classes. Perfect predictions of visual features would result in well-aligned crosses and circles of the same color. Plots for \textbf{CUB} and \textbf{SUN} are based on their first splits. Plots for \textbf{ImageNet} are based on randomly selected 48 unseen classes from \emph{2-hop} and word vectors as  semantic representations. Best viewed in color. See the supplementary material for larger figures.} \label{fTsne}
\vspace{-5pt}
\end{figure*}

We first show that predicted visual exemplars better reflect visual similarities between classes than semantic representations.
Let $\mD_{\va_u}$ be the pairwise Euclidean distance matrix between \emph{unseen} classes computed from semantic representations (i.e., $\cU$ by $\cU$), $\mD_{\vpsi(\va_u)}$ the distance matrix computed from predicted exemplars, and $\mD_{\vv_u}$ the distance matrix computed from real exemplars (which we do not have access to).
Table~\ref{tCorrDis} shows that the correlation between $\mD_{\vpsi(\va_u)}$ and $\mD_{\vv_u}$ is much higher than that between $\mD_{\va_u}$ and $\mD_{\vv_u}$. Importantly, we improve this correlation without access to any data of the unseen classes. See also similar results using another metric in the supplementary material.

We then show some t-SNE~\cite{tSNE} visualization of predicted visual exemplars of the \emph{unseen} classes. 
Ideally, we would like them to be as close to their corresponding real images as possible. In Fig.~\ref{fTsne}, we demonstrate that this is indeed the case for many of the unseen classes; for those unseen classes (each of which denoted by a color), their real images (crosses) and our predicted visual exemplars (circles) are well-aligned.

The quality of predicted exemplars (in this case based on the distance to the real images) depends on two main factors: the predictive capability of semantic representations and the number of semantic representation-visual exemplar pairs available for training, which in this case is equal to the number of seen classes $\cS$. On \textbf{AwA} where we have only $40$ training pairs, the predicted exemplars are surprisingly accurate, mostly either placed in their corresponding clusters or at least closer to their clusters than predicted exemplars of the other unseen classes. Thus, we expect them to be useful for discriminating among the unseen classes. On \textbf{ImageNet}, the predicted exemplars are not as accurate as we would have hoped, but this is expected since the word vectors are purely learned from text. 

We also observe relatively well-separated clusters in the semantic embedding space (in our case, also the visual feature space since we only apply PCA projections to the visual features), confirming our assumption about the existence of clustering structures.
On \textbf{CUB}, we observe that these clusters are more mixed than on other datasets. This is not surprising given that it is a fine-grained classification dataset of bird species.

\subsection{Zero-shot learning results}

\subsubsection{Main results}

\begin{table}
\centering
\caption{\small Comparison between existing ZSL approaches in multi-way classification accuracies (in \%) on four benchmark datasets. For each dataset, we mark the best in red and the second best in blue. \emph{Italic} numbers denote \emph{per-sample} accuracy instead of \emph{per-class} accuracy. On \textbf{ImageNet}, we report results for both types of semantic representations: Word vectors (wv) and MDS embeddings derived from WordNet (hie). All the results are based on GoogLeNet features \cite{SzegedyLJSRAEVR14}.}. \label{tbMain}
\vskip -0.5em
\footnotesize
\begin{tabular}{c|c|c|c|c|c}
\text{Approach} & \textbf{AwA} & \textbf{CUB} & \textbf{SUN} & \multicolumn{2}{|c}{\textbf{ImageNet}}\\ \cline{5-6}
	 &  & &  	 & wv & hie\\ \hline
\mtt{ConSE}$^\dagger$ \cite{NorouziMBSSFCD14}  & 63.3 & 36.2 & 51.9 & \emph{1.3} &-\\
\mtt{BiDiLEL} \cite{WangC16}  & 72.4 & 49.7$^\S$ & - & - & -\\
\mtt{LatEm}$^\ddagger$ \cite{XianASNHS16}  & 72.1 & 48.0 & 64.5 & - & -\\
\mtt{CCA} \cite{Lu16}  & - & - & - & - & \emph{1.8}\\
\mtt{SynC}$^\textrm{o-vs-o}$ \cite{ChangpinyoCGS16}  & 69.7 & 53.4& 62.8 & \emph{1.4} & {\color{red}\textbf{\emph{2.0}}}\\
\mtt{SynC}$^\textrm{struct}$ \cite{ChangpinyoCGS16}  & 72.9 	& 54.5 	& 62.7 & \emph{1.5} & -\\ \hline
\mtt{EXEM (ConSE)}  & 70.5 & 46.2  & 60.0 & - & -\\
\mtt{EXEM (LatEm)}$^\ddagger$ & 72.9 & 56.2  & {\color{blue}\textbf{67.4}} & - & -\\
\mtt{EXEM (SynC$^\textrm{o-vs-o}$)}  & 73.8 & 56.2  & 66.5 & \emph{1.6} & {\color{red}\textbf{\emph{2.0}}}\\
\mtt{EXEM (SynC$^\textrm{struct}$)}  & {\color{red}\textbf{77.2}} & {\color{red}\textbf{59.8}} & 66.1 & - & -\\
\mtt{EXEM (1NN)}  & 76.2	& 56.3 	& {\color{red}\textbf{69.6}} & {\color{blue}\textbf{\emph{1.7}}} & {\color{red}\textbf{\emph{2.0}}}\\
\mtt{EXEM (1NNs)}  & {\color{blue}\textbf{76.5}} 	& {\color{blue}\textbf{58.5}}	& {\color{blue}\textbf{67.3}} &  {\color{red}\textbf{\emph{1.8}}} & {\color{red}\textbf{\emph{2.0}}}\\ \hline
\end{tabular}
\vskip -1.0em
\begin{flushleft}
$^\S$: on a particular split of seen/unseen classes. \hspace{13pt} $^\dagger$: reported in~\cite{ChangpinyoCGS16}.\\
$^\ddagger$: based on the code of~\cite{XianASNHS16}, averaged over 5 different initializations.

\end{flushleft}
\vskip -1.5em
\end{table}

Table~\ref{tbMain} summarizes our results in the form of multi-way classification accuracies on all datasets.
We significantly outperform recent state-of-the-art baselines when using GoogLeNet features. In the supplementary material, we provide additional quantitative and qualitative results, including those on generalized zero-shot learning task~\cite{ChaoCGS16}. 

We note that, on \textbf{AwA}, several recent methods obtain higher accuracies due to using a more optimistic evaluation metric (per-sample accuracy) and new types of deep features~\cite{ZhangXG16,ZhangS15}. This has been shown to be unsuccessfully replicated (cf. Table~2 in ~\cite{XianAS17}). See the supplementary material for results of these and other less competitive baselines.

Our alternative approach of treating predicted visual exemplars as the ideal semantic representations significantly outperforms taking semantic representations as given. \mt{EXEM (SynC)}, \mt{EXEM (ConSE)}, \mt{EXEM (LatEm)} outperform their corresponding \emph{base} ZSL methods relatively by 5.9-6.8\%, 11.4-27.6\%, and 1.1-17.1\%, respectively. This again suggests improved quality of semantic representations (on the predicted exemplar space).

Furthermore, we find that there is no clear winner between using predicted exemplars as ideal semantic representations or as data prototypes. The former seems to perform better on datasets with fewer seen classes. Nonetheless, we remind that using 1-nearest-neighbor classifiers clearly scales much better than zero-shot learning methods; \mt{EXEM (1NN)} and \mt{EXEM (1NNs)} are more efficient than \mt{EXEM (SynC)}, \mt{EXEM (ConSE)}, and \mt{EXEM (LatEm)}.

Finally, we find that in general using the standardized Euclidean distance instead of the Euclidean distance for nearest neighbor classifiers helps improve the accuracy, especially on \textbf{CUB}, suggesting there is a certain effect of collapsing actual data during training. The only exception is on \textbf{SUN}. We suspect that the standard deviation values computed on the seen classes on this dataset may not be robust enough as each class has only 20 images.

\subsubsection{Large-scale zero-shot classification results}

\begin{table*}
\centering
\caption{\small Comparison between existing ZSL approaches on \textbf{ImageNet} using \textbf{word vectors} of the class names as semantic representations. For both metrics (in \%), the higher the better. The best is in red. The numbers of unseen classes are listed in parentheses. $^\dagger$: reported in~\cite{ChangpinyoCGS16}.} 
\label{tbImagenet}
\footnotesize
\begin{tabular}{c|c|ccccc|cccc}
\text{Test data} & \text{Approach} & \multicolumn{5}{|c|}{Flat Hit@K} & \multicolumn{4}{|c}{Hierarchical precision@K}\\ \cline{3-11}
& K= & \text{1} & \text{2} & \text{5} & \text{10} & \text{20} & \text{2} & \text{5} & \text{10} & \text{20} \\ \hline
& \mtt{ConSE}$^\dagger$~\cite{NorouziMBSSFCD14}\hspace{4pt} & 8.3 & 12.9 & 21.8 & 30.9 & 41.7 & 21.5 & 23.8 & 27.5 & 31.3 \\ 
& \mtt{SynC}$^\textrm{o-vs-o}$ \cite{ChangpinyoCGS16} & 10.5 & 16.7 & 28.6 & 40.1 & 52.0 & 25.1 & 27.7 & 30.3 & 32.1 \\ \cline{2-11}
\emph{2-hop}

(1,509)
& \mtt{EXEM (SynC$^\textrm{o-vs-o}$)} & 11.8 & 18.9 & 31.8 & 43.2 & 54.8 & 25.6 & 28.1 & 30.2 & 31.6 \\
& \mtt{EXEM (1NN)}	& 11.7 & 18.3 & 30.9 & 42.7 & 54.8 & 25.9 & 28.5 & {\color{red}\textbf{31.2}} &{\color{red}\textbf{33.3}} \\
& \mtt{EXEM (1NNs)} & {\color{red}\textbf{12.5}} & {\color{red}\textbf{19.5}} & {\color{red}\textbf{32.3}} & {\color{red}\textbf{43.7}} & {\color{red}\textbf{55.2}} & {\color{red}\textbf{26.9}} & {\color{red}\textbf{29.1}} & 31.1 & 32.0 \\
\hline
& \mtt{ConSE}$^\dagger$~\cite{NorouziMBSSFCD14}\hspace{4pt} & 2.6 & 4.1 & 7.3 & 11.1 & 16.4 & 6.7 & 21.4 & 23.8 & 26.3 \\ 
& \mtt{SynC}$^\textrm{o-vs-o}$ \cite{ChangpinyoCGS16} & 2.9 & 4.9 & 9.2 & 14.2 & 20.9 & 7.4 & 23.7 & 26.4 & 28.6 \\ \cline{2-11}
\emph{3-hop} 
(7,678)
& \mtt{EXEM (SynC$^\textrm{o-vs-o}$)}  & 3.4 & 5.6 & 10.3 & 15.7 & 22.8 & 7.5 & 24.7 & 27.3 & 29.5 \\ 
& \mtt{EXEM (1NN)}& 3.4 & 5.7 & 10.3 & 15.6 & 22.7 & 8.1 & {\color{red}\textbf{25.3}} & {\color{red}\textbf{27.8}} & {\color{red}\textbf{30.1}}	 \\
& \mtt{EXEM (1NNs)} & {\color{red}\textbf{3.6}} & {\color{red}\textbf{5.9}} & {\color{red}\textbf{10.7}} & {\color{red}\textbf{16.1}} & {\color{red}\textbf{23.1}} & {\color{red}\textbf{8.2}} & 25.2 & 27.7 & 29.9 \\
\hline
& \mtt{ConSE}$^\dagger$~\cite{NorouziMBSSFCD14}\hspace{4pt} & 1.3 & 2.1 & 3.8 & 5.8 & 8.7 & 3.2 & 9.2 & 10.7 & 12.0 \\ 
& \mtt{SynC}$^\textrm{o-vs-o}$ \cite{ChangpinyoCGS16} & 1.4 & 2.4 & 4.5 & 7.1 & 10.9 & 3.1 & 9.0 & 10.9 & 12.5 \\ \cline{2-11}
\emph{All}
(20,345)
& \mtt{EXEM (SynC$^\textrm{o-vs-o}$)} & 1.6 & 2.7 & 5.0 & 7.8 & 11.8 & 3.2 & 9.3 & 11.0 & 12.5 \\ 
& \mtt{EXEM (1NN)}& 1.7 & 2.8 & 5.2 & 8.1 & 12.1 & {\color{red}\textbf{3.7}} & {\color{red}\textbf{10.4}} & {\color{red}\textbf{12.1}} & {\color{red}\textbf{13.5}} \\
& \mtt{EXEM (1NNs)}  & {\color{red}\textbf{1.8}} & {\color{red}\textbf{2.9}} & {\color{red}\textbf{5.3}} & {\color{red}\textbf{8.2}} & {\color{red}\textbf{12.2}} & 3.6 & 10.2 & 11.8 & 13.2\\
\hline
\end{tabular}
\vskip -1em
\end{table*}

\begin{table}
\centering
\caption{\small Comparison between existing ZSL approaches on \textbf{ImageNet} (with 20,842 unseen classes) using \textbf{MDS embeddings derived from WordNet \cite{Lu16}} as semantic representations. The higher, the better (in \%). The best is in red.} 
\label{tbImagenetMDS}
\footnotesize
\begin{tabular}{c|c|ccccc}
\text{Test data} & \text{Approach} & \multicolumn{5}{|c}{Flat Hit@K} \\ \cline{3-7}
& K= & \text{1} & \text{2} & \text{5} & \text{10} & \text{20} \\ \hline 
& \mtt{CCA}~\cite{Lu16}\hspace{4pt} & 1.8 & 3.0 & 5.2 & 7.3 & 9.7  \\ 
\emph{All}
& \mtt{SynC}$^\textrm{o-vs-o}$ \cite{ChangpinyoCGS16}& {\color{red}\textbf{2.0}} & 3.4 & 6.0 & 8.8 & 12.5  \\ \cline{2-7}
(20,842)
& \mtt{EXEM (SynC$^\textrm{o-vs-o}$)} & {\color{red}\textbf{2.0}} & 3.3 & 6.1 & 9.0 & 12.9 \\
& \mtt{EXEM (1NN)}	& {\color{red}\textbf{2.0}} & {\color{red}\textbf{3.4}} & {\color{red}\textbf{6.3}} & {\color{red}\textbf{9.2}} & 13.1  \\
& \mtt{EXEM (1NNs)} & {\color{red}\textbf{2.0}} & {\color{red}\textbf{3.4}} & 6.2 & {\color{red}\textbf{9.2}} & {\color{red}\textbf{13.2}} \\
\hline
\end{tabular}
\end{table}

We then provide expanded results for \textbf{ImageNet}, following evaluation protocols in the literature. 
In Table~\ref{tbImagenet} and~\ref{tbImagenetMDS}, we provide results based on the exemplars predicted by word vectors and MDS features derived from WordNet, respectively. We consider 
\mt{SynC}$^\textrm{o-v-o}$, rather than \mt{SynC}$^\textrm{struct}$, as the former shows better performance on \textbf{ImageNet}~\cite{ChangpinyoCGS16}.
Regardless of the types of metrics used, our approach outperforms the baselines significantly when using word vectors as semantic representations. For example, on \emph{2-hop}, we are able to improve the F@1 accuracy by 2\% over the state-of-the-art. However, we note that this improvement is not as significant when using MDS-WordNet features as semantic representations.

We observe that the 1-nearest-neighbor classifiers perform better than using predicted exemplars as more powerful semantic representations. We suspect that, when the number of classes is very high, zero-shot learning methods (\mt{ConSE} or \mt{SynC}) do not fully take advantage of the \emph{meaning} provided by each dimension of the exemplars.

\subsubsection{From zero-shot to few-shot learning}

\begin{figure}[t]
\centering
\includegraphics[width=0.4\textwidth]{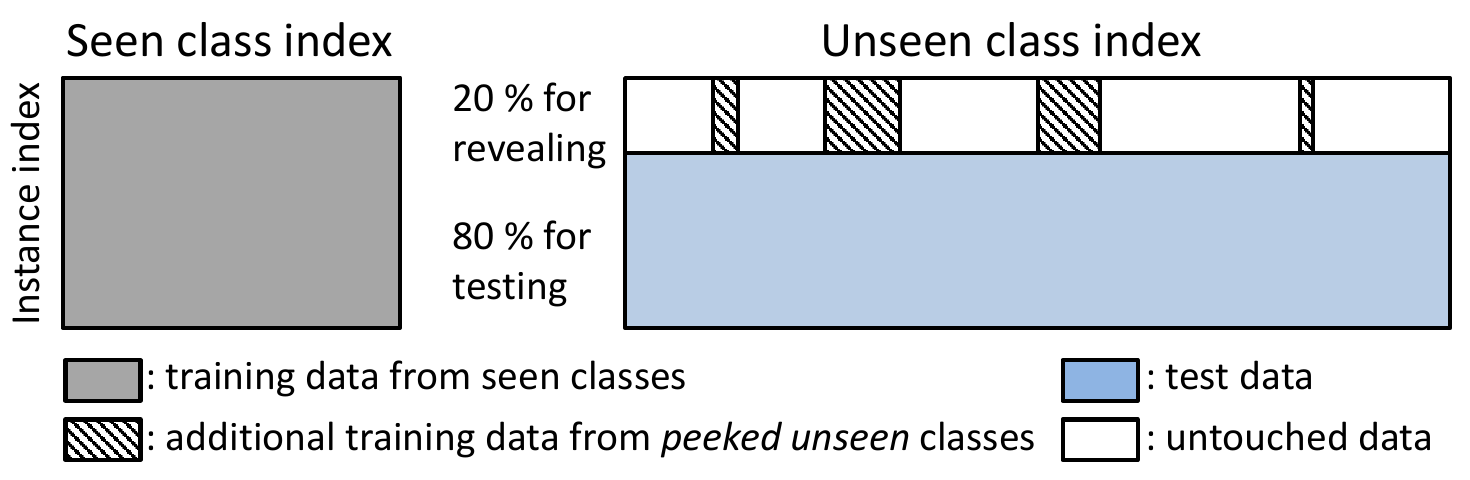}
\vspace{-2pt}
\caption{\small Data split for zero-to-few-shot learning on \textbf{ImageNet}} \label{fHybridSetting}
\vspace{-12pt}
\end{figure}

In this section, we investigate what will happen when we allow ZSL algorithms to \emph{peek} into some labeled data from part of the unseen classes. 
Our focus will be on \emph{All} categories of \textbf{ImageNet}, two ZSL methods (\mt{SynC}$^\textrm{o-vs-o}$ and \mt{EXEM (1NN)}), and two evaluation metrics (F@1 and F@20). For brevity, we will denote \mt{SynC}$^\textrm{o-vs-o}$ and \mt{EXEM (1NN)} by \mt{SynC} and \mt{EXEM}, respectively.\\[0.25em]
\noindent\textbf{Setup} 
We divide images from each unseen class into two sets.
The first 20\% are reserved as training examples that may or may not be revealed. 
This corresponds to on average 127 images per class.
If revealed, those \emph{peeked} unseen classes will be marked as seen, and their labeled data can be used for training.
The other 80\% are for testing. The test set is always fixed such that we have to do few-shot learning for peeked unseen classes and zero-shot learning on the rest of the unseen classes.
Fig.~\ref{fHybridSetting} summarizes this protocol.

We then vary the number of peeked unseen classes $B$. Also, for each of these numbers, we explore the following subset selection strategies (more details are in the supplementary material): (i) \textbf{Uniform random}: Randomly selected $B$ unseen classes from the uniform distribution; (ii) \textbf{Heavy-toward-seen random} Randomly selected $B$ classes that are semantically similar to seen classes according to the WordNet hierarchy; (iii) \textbf{Light-toward-seen random} Randomly selected $B$ classes that are semantically far away from seen classes; (iv) \textbf{K-means clustering for coverage} Classes whose semantic representations are nearest to each cluster's center, where semantic embeddings of the unseen classes are grouped by k-means clustering with k $= B$; (v) \textbf{DPP for diversity} Sequentially selected classes by a greedy algorithm for fixed-sized determinantal point processes (k-DPPs) \cite{KuleszaT11} with the RBF kernel computed on semantic representations.

\begin{figure}[t]
\centering
\includegraphics[width=0.4\textwidth]{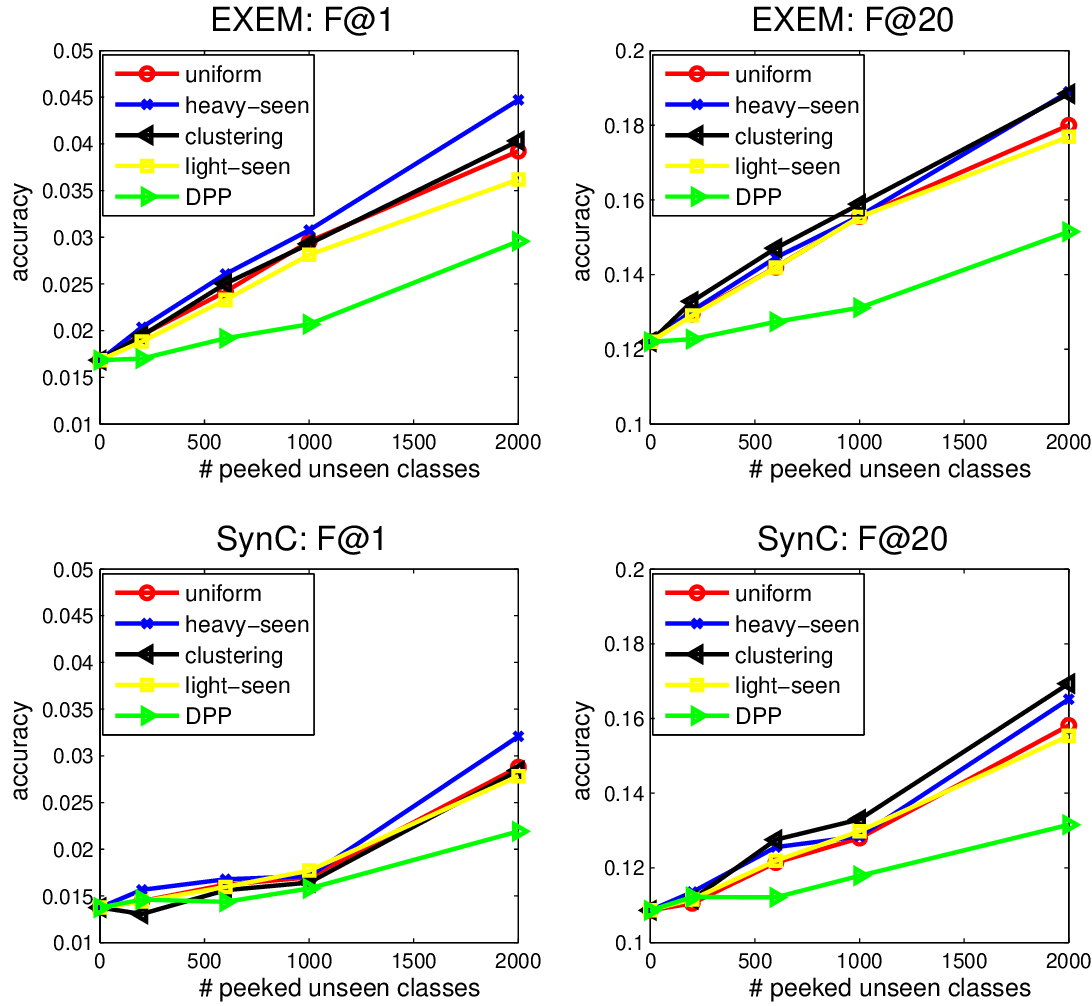}
\vspace{-5pt}
\caption{\small Accuracy vs. the number of peeked unseen classes for \mtc{EXEM} (top) and \mtc{SynC} (bottom) \textbf{across different subset selection methods}. Evaluation metrics are F@1 (left) and F@20 (right).} \label{fCompareSynCSelect}
\vspace{-5pt}
\end{figure}

\noindent\textbf{Results} For each of the ZSL methods (\mt{EXEM} and \mt{SynC}), we first compare different subset selection methods when the number of peeked unseen classes is small (up to 2,000) in Fig.~\ref{fCompareSynCSelect}. We see that the performances of different subset selection methods are consistent across ZSL methods. Moreover, \emph{heavy-toward-seen classes} are preferred for \emph{strict} metrics (Flat Hit@1) but \emph{clustering} is preferred for \emph{flexible} metrics (Flat Hit@20).
This suggests that, for a strict metric, it is better to \emph{pick the classes that are semantically similar to what we have seen}. On the other hand, if the metric is flexible, we should focus on providing \emph{coverage} for all the classes so each of them has knowledge they can transfer from.

\begin{figure}[t]
\centering
\includegraphics[width=0.4\textwidth]{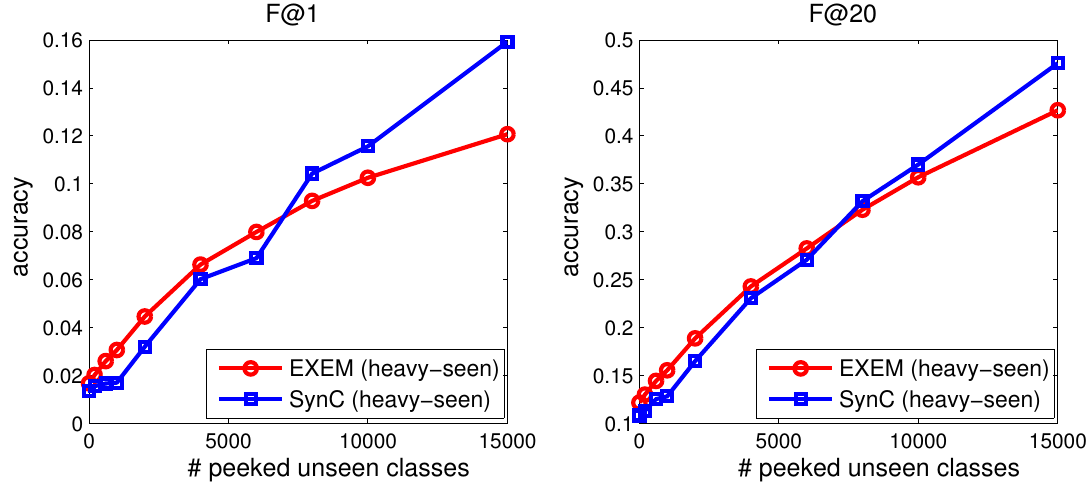}
\vspace{-2pt}
\caption{\small Accuracy vs. the number of peeked unseen classes for \mtc{EXEM} and \mtc{SynC} for heavy-toward-seen class selection strategy. Evaluation metrics are F@1 (left) and F@20 (right).} \label{fCompareSynC}
\vspace{-8pt}
\end{figure}

Next, using the best performing heavy-toward-seen selection, we focus on comparing \mt{EXEM} and \mt{SynC} with larger numbers of peeked unseen classes in Fig.~\ref{fCompareSynC}. When the number of peeked unseen classes is small, \mt{EXEM} outperforms \mt{SynC}. (In fact, \mt{EXEM} outperforms \mt{SynC} for \emph{each} subset selection method in Fig.~\ref{fCompareSynCSelect}.) However, we observe that \mt{SynC} will finally catch up and surpass \mt{EXEM}. This is not surprising; as we observe more labeled data (due to the increase in peeked unseen set size), the setting will become more similar to supervised learning (few-shot learning), where linear classifiers used in \mt{SynC} should outperform nearest center classifiers used by \mt{EXEM}.
Nonetheless, we note that \mt{EXEM} is more computationally advantageous than \mt{SynC}. In particular, when training on 1K classes of \textbf{ImageNet} with over 1M images, \mt{EXEM} takes 3 mins while \mt{SynC} 1 hour.
We provide additional results under this scenario in the supplementary material.

\subsubsection{Analysis}
\label{sec_analysis_svr_pca}
\noindent\textbf{PCA or not?}
Table~\ref{tPCA} investigates the effect of PCA. In general, \mt{EXEM (1NN)} performs comparably with and without PCA.
Moreover, decreasing PCA projected dimension $\cst{d}$ from 1024 to 500 does not hurt the performance.
Clearly, a smaller PCA dimension leads to faster computation due to fewer regressors to be trained. See additional results with other values for $\cst{d}$ in the supplementary material.

\begin{table}
\centering
{\footnotesize
\caption{\small Accuracy of \mtt{EXEM (1NN)} on \textbf{AwA}, \textbf{CUB}, and \textbf{SUN} when predicted exemplars are from original visual features (No PCA) and PCA-projected features (PCA with $\cst{d}$ = 1024 and $\cst{d}$ = 500).}
\vskip .5em
\label{tPCA}
\begin{tabular}{c|c|c|c}
Dataset & No PCA & PCA & PCA  \\ 
name & $\cst{d}$ = 1024  & $\cst{d}$ = 1024 & $\cst{d}$ = 500\\ \hline
\textbf{AwA} & \textbf{77.8} & 76.2 & 76.2\\ \hline
\textbf{CUB} & 55.1 & \textbf{56.3} & \textbf{56.3}\\ \hline
\textbf{SUN} & 69.2 & \textbf{69.6} & \textbf{69.6}\\ \hline
\end{tabular}
}
\vskip -8pt
\end{table}

\begin{table}
\centering
{\footnotesize
\caption{\small Comparison between \mtt{EXEM (1NN)} with support vector regressors (SVR) and with 2-layer multi-layer perceptron (MLP) for predicting visual exemplars. Results on \textbf{CUB} are for the first split. Each number for MLP is an average over 3 random initialization.}
\vskip .5em
\label{tSVRvsMLP}
\begin{tabular}{c|c|c|c|c}
Dataset & How to predict & No PCA & PCA & PCA  \\ 
name &  exemplars &$\cst{d}$ = 1024  & $\cst{d}$ = 1024 & $\cst{d}$ = 500\\ \hline
\textbf{AwA} & SVR & \textbf{77.8} & 76.2 & \textbf{76.2}\\ \cline{2-5}
						 & MLP & 76.1 $\pm$ 0.5 & \textbf{76.4} $\pm$ 0.1 & 75.5 $\pm$ 1.7\\ \hline
\textbf{CUB} & SVR & \textbf{57.1} & \textbf{59.4} & \textbf{59.4} \\ \cline{2-5}
						 & MLP & 53.8 $\pm$ 0.3 & 54.2 $\pm$ 0.3 & 53.8 $\pm$ 0.5\\ \hline
\end{tabular}
}
\vskip -1em
\end{table}

\noindent\textbf{Kernel regression vs. Multi-layer perceptron} We compare two approaches for predicting visual exemplars: kernel-based support vector regressors (SVR) and 2-layer multi-layer perceptron (MLP) with ReLU nonlinearity.  
MLP weights are $\ell_2$ regularized, and we cross-validate the regularization constant.
Additional details are in the supplementary material.

Table~\ref{tSVRvsMLP} shows that SVR performs more robustly than MLP.
One explanation is that MLP is prone to overfitting due to the small training set size (the number of seen classes) as well as the model selection challenge imposed by ZSL scenarios.
SVR also comes with other benefits; it is more efficient and less susceptible to initialization.

\section{Related Work}
\label{sRelated}

ZSL has been a popular research topic in both computer vision and machine learning. A general theme is to make use of semantic representations such as attributes or word vectors to relate visual features of the seen and unseen classes, as summarized in~\cite{AkataPHS13}.

Our approach for predicting visual exemplars is inspired by~\cite{FromeCSBDRM13,NorouziMBSSFCD14}. They predict an image's semantic embedding from its visual features and compare to unseen classes' semantic embeddings. As mentioned in Sect.~\ref{sEXEMcompare}, we perform ``inverse prediction'': given an unseen class's semantic representation, we predict where the exemplar visual feature vector for that class is in the semantic embedding space.

There has been a recent surge of interest in applying deep learning models to generate images~\cite{MansimovPBS16,ReedAYLLS16,YanYSL16}. Most of these methods are based on probabilistic models (in order to incorporate the statistics of natural images). Unlike them, our prediction is to purely deterministically predict visual exemplars (features).  Note that, generating features directly is likely easier and more effective than generating realistic images first and then extracting visual features from them.

\section{Discussion}
\label{sDiscuss}
We have proposed a novel ZSL model that is simple but very effective. Unlike previous approaches, our method directly solves ZSL by predicting visual exemplars --- cluster centers that characterize visual features of the unseen classes of interest. This is made possible partly due to the well separate cluster structure in the deep visual feature space. 
We apply predicted exemplars to the task of zero-shot classification based on two views of these exemplars: ideal semantic representations and prototypical data points. 
Our approach achieves state-of-the-art performance on multiple standard benchmark datasets. Finally, we also analyze our approach and compliment our empirical studies with an extension of zero-shot to few-shot learning.
\vskip .5em
\textbf{Acknowledgements} {\small This work is partially supported by USC Graduate Fellowship, NSF IIS-1065243, 1451412, 1513966/1632803, 1208500, CCF-1139148, a Google Research Award, an Alfred. P. Sloan Research Fellowship and ARO\# W911NF-12-1-0241 and W911NF-15-1-0484.}
{\small
\bibliographystyle{ieee}
\bibliography{main_iccv}
}

\clearpage

{
\begin{center}
{\Large \bf Supplementary Material:\\Predicting Visual Exemplars of Unseen Classes for Zero-Shot Learning \par}
\end{center}
}
\appendix

This supplementary material provides the following details omitted in the main text.
\begin{itemize}
\item Sect.~\ref{sSupplApproach}: Details on our proposed zero-shot learning method (Sect. 2.1 of the main text)
\item Sect.~\ref{sSupplSetup}: Details on the experimental setup, including details on datasets, details on baselines, and hyper-parameter tuning (Sect. 3.1 of the main text)
\item Sect.~\ref{sSupplPredExemTop}: Expanded and additional results on the predicted exemplars, including another metric for evaluating the quality of predicted exemplars, and larger visualization (Sect. 3.2 of the main text).
\item Sect.~\ref{sSupplZSLexp}: Additional experimental results on ZSL, including expanded Table 3 and Table 4, ZSL with word vectors as semantic representations, and qualitative results (Sect. 3.3.1 and 3.3.2 of the main text). 
\item Sect.~\ref{sSupplGZSLexp}: Generalized zero-shot learning results
\item Sect.~\ref{sSupplPeeked}: Details and additional results on zero-shot to few-shot learning experiments, including details on how to select a subset of peeked unseen classes. (Sect. 3.3.3 of the main text)
\item Sect.~\ref{sSupplPCA}: Additional analysis on dimension for PCA  (Sect. 3.3.4 of the main text)
\item Sect.~\ref{sSupplMLP}: Details on multi-layer perceptron (Sect. 3.3.4 of the main text)
\end{itemize}

\section{Details on our proposed zero-shot learning method}
\label{sSupplApproach}

\paragraph{SVR formulation for predicting visual exemplars}

In Sect.~2.1 of the main text, given semantic representation-visual exemplar pairs of the seen classes, we learn $\cst{d}$ support vector regressors (SVR) with RBF kernel. Specifically, for each dimension $d = 1, \ldots, \cst{d}$ of $\vv_{c}$, SVR is learned based on the $\nu$-SVR formulation \cite{ScholkopfSWB00}:
\begin{align}
\min_{\vw, \xi, \xi', \epsilon} \frac{1}{2} & \vw^T\vw + \lambda(\nu \epsilon + \frac{1}{\cS} \sum_{c = 1}^{\cS} (\xi_c + \xi'_c)) \nonumber\\
\mathsf{s.t.} & \vw^T \vtheta^{\mathsf{rbf}}(\va_c)  -  \vv_{c} \leq \epsilon + \xi_c \\
& \vv_{c} - \vw^T \vtheta^{\mathsf{rbf}}(\va_c)  \leq \epsilon + \xi'_c \nonumber\\
& \xi_c \geq 0, \xi'_c \geq 0, \nonumber
\end{align}
where $\vtheta^{\mathsf{rbf}}$ is an implicit nonlinear mapping based on our kernel. We have dropped the subscript $d$ for aesthetic reasons but readers are reminded that each regressor is trained independently with its own target values (i.e., ${\vv_{c}}_{d}$) and parameters (i.e., $\vw_d$). We found that the regression error is not sensitive to $\lambda$ and set it to 1 in all experiments except for zero-shot to few-shot learning. We jointly tune $\nu \in (0,1]$ and the kernel bandwidth and finally apply the same set of hyper-parameters for all the $\cst{d}$ regressors. Details on hyper-parameter tuning can be found in Sect.~\ref{sSupplHPTuning}. The resulting $\vpsi(\cdot) = [\vw_1^{T}\vtheta^{\mathsf{rbf}}(\cdot), \cdots, \vw_\cst{d}^{T}\vtheta^{\mathsf{rbf}}(\cdot)]^T$, where $\vw_d$ is from the $d$-th regressor.

\section{Details on the experimental setup}
\label{sSupplSetup}

\subsection{Additional information on datasets}
We experiment on four benchmark datasets. The \textbf{Animals with Attributes (AwA)} dataset~\cite{LampertNH14} consists of 30,475 images of 50 animal classes, along with a standard data split for zero-shot learning --- 40 seen classes (for training) and 10 unseen classes. 
The \textbf{CUB-200-2011 Birds (CUB)}~\cite{WahCUB_200_2011} has 200 bird classes and 11,788 images, while the \textbf{SUN Attribute (SUN)} dataset~\cite{PattersonH14} contains 14,340 images of 717 scene categories (20 images from each category). We follow seen/unseen splits in ~\cite{LampertNH14} for \textbf{AwA}, and \cite{ChangpinyoCGS16} for \textbf{CUB} (4 splits) and \textbf{SUN} (10 splits). We report average results from all the splits.

On \textbf{ImageNet}~\cite{deng2009imagenet}, we follow the setting in \cite{FromeCSBDRM13, NorouziMBSSFCD14,ChangpinyoCGS16}. 
We use the ILSVRC 2012 1K dataset~\cite{ILSVRC15}, which contains 1,281,167 training and 50,000 validation images from 1,000 categories, as data from seen classes.
Images of unseen classes come from the rest of the ImageNet Fall 2011 release dataset~\cite{deng2009imagenet} that do not overlap with any of those 1,000 categories. 
In total, this dataset consists of 14,197,122 images from 21,841 classes, \textbf{20,842 unseen classes} of which are unseen ones. Note that, as mentioned in \cite{ChangpinyoCGS16}, there is one class in the ILSVRC 2012 1K dataset that does not appear in the ImageNet 2011 21K dataset. Thus, we have a total of 20,842 unseen classes to evaluate.

\subsection{Details on ZSL baselines}
\label{sSuppZSLBaselines}

We focus on comparing our method with a recent state-of-the-art baseline \mt{SynC}~\cite{ChangpinyoCGS16}. Specifically, we adopt the version that sets the number of base classifiers to be $\cS$ (the number of seen classes), and sets $\vb_r = \va_c$ for $r=c$ (cf. Sec. 2.2.2 of the main text). Note that this is the version that has reported results on all four datasets.

\mt{SynC} has been shown to outperform multiple strong baselines under the same setting. In particular, under the setting of \cite{ChangpinyoCGS16} which we adopt in this paper, \mt{SynC} outperforms \mt{SJE} \cite{AkataRWLS15}, \mt{ESZSL} \cite{Bernardino15}, \mt{COSTA} \cite{MensinkGS14}, and \mt{ConSE} \cite{NorouziMBSSFCD14}. For more details, see Table 3 and Table 4 in \cite{ChangpinyoCGS16}. Under the setting of \cite{XianAS17} (with ResNet deep features \cite{HeZRS16} and standard dataset splits), \mt{SynC} is the best performing method among \cite{LampertNH14,AkataPHS13,FromeCSBDRM13,SocherGMN13,NorouziMBSSFCD14,AkataRWLS15,Bernardino15,ZhangS15,XianASNHS16} on average on \textbf{AwA}, \textbf{CUB}, \textbf{SUN}, and additionally \textbf{aPY} \cite{FarhadiEHF09} (See Fig.~1 in \cite{XianAS17}), and by far the best performing method on \textbf{ImageNet} among \cite{AkataPHS13,FromeCSBDRM13,SocherGMN13,NorouziMBSSFCD14,AkataRWLS15,Bernardino15,XianASNHS16} (See Table 4 in \cite{XianAS17}). Note that Xian et al. \cite{XianAS17} recently proposes alternative splits of the datasets. In some scenarios, \mt{SynC} may not perform best on these splits. We leave further investigation of the performance of our ZSL method on these newly proposed splits for future work. 

Besides \mt{SynC}, in Table 3 and Table 5 of the main text, we also include ZSL recognition accuracies of \emph{recent} ZSL methods that have not been compared in \cite{ChangpinyoCGS16}, including \mt{BiDiLEL} \cite{WangC16}, \mt{LatEm} \cite{XianASNHS16}, and \mt{CCA} \cite{Lu16}. For each of these methods, we strive to ensure fair comparison in terms of semantic representations, visual features, and evaluation metrics.

\subsection{Hyper-parameter tuning}
\label{sSupplHPTuning}
There are several hyper-parameters to be tuned in our experiments: \textbf{(a)} projected dimensionality $\cst{d}$ for PCA and \textbf{(b)} $\lambda$, $\nu$, and the RBF-kernel bandwidth in SVR. 
For \textbf{(a)}, we found that the ZSL performance is not sensitive to $\cst{d}$ and thus set $\cst{d}=500$ for all experiments. 
For \textbf{(b)}, we perform \emph{class-wise} cross-validation (CV), following previous work~\cite{ChangpinyoCGS16, ElhoseinySE13, ZhangS15}, with two exceptions. First, we found $\lambda=1$ works robustly on all datasets for zero-shot learning. Second, we fix all the hyper-parameters when we increase the number of peeked unseen classes (c.f. Sect. 3.3.3 of the main text) in the case of \mt{EXEM (1NN)}\footnote{In the experiments where we peek into some unseen classes' examples, we find that, for \mtt{EXEM (1NN)}, fixing the hyper-parameters tuned on ZSL (with 0 peeked unseen classes) works robustly for other numbers of peeked unseen classes. However, this is not the case for \mtt{SynC}, in which case we tune the hyper-parameters for different numbers of peeked unseen classes.}.
 
The \emph{class-wise} CV can be done as follows. We hold out data from a subset of seen classes as pseudo-unseen classes, train our models on the remaining folds (which belong to the remaining classes), and tune hyper-parameters based on a certain performance metric on the held-out fold.
This scenario simulates the ZSL setting and has been shown to outperform the conventional CV in which each fold contains a portion of training examples from all classes~\cite{ChangpinyoCGS16}.

We consider the following two performance metrics.
The first one minimizes the distance between the predicted exemplars and the ground-truth (average of PCA-projected validation data of each class) in $\R^{\cst{d}}$.
We use the Euclidean distance in this case. We term this measure \textbf{CV-distance}. This approach does not assume the downstream task at training and aims to measure the quality of predicted exemplars by its \emph{faithfulness}. 

The other approach maximizes the zero-shot classification accuracy on the validation set. 
This measure can easily be obtained for \mt{EXEM (1NN)} and \mt{EXEM (1NNs)}, which use simple decision rules that have no further hyper-parameters to tune. Empirically, we found that \textbf{CV-accuracy} generally leads to slightly better performance. The results reported in the main text for these two approaches are thus based on this measure. 

On the other hand, \mt{EXEM (SynC$^{\textrm{o-vs-o}}$)}, \mt{EXEM (SynC$^{\textrm{struct}}$)}, \mt{EXEM (ConSE)}, and \mt{EXEM (LatEm)} require further hyper-parameter tuning. For computational purposes, we use \textbf{CV-distance} for tuning hyper-parameters of the regressors, followed by the hyper-parameter tuning for \mt{SynC} and \mt{ConSE} using the predicted exemplars. Since \mt{SynC} and \mt{ConSE} construct their classifiers based on the distance values between class semantic representations, we do not expect a significant performance drop in this case. (We remind the reader that, in \mt{EXEM (SynC$^{\textrm{o-vs-o}}$)}, \mt{EXEM (SynC$^{\textrm{struct}}$)}, \mt{EXEM (ConSE)}, and \mt{EXEM (LatEm)}, the predicted exemplars are used as semantic representations.)

\section{Expanded and additional results on the predicted exemplars}
\label{sSupplPredExemTop}

\subsection{Another metric for evaluating the quality of predicted visual exemplars}

\begin{table}
\centering
\caption{\small Overlap of k-nearest classes (in \%) on \textbf{AwA}, \textbf{CUB}, \textbf{SUN}. We measure the overlap between those searched by real exemplars and those searched by semantic representations (i.e., attributes) or predicted exemplars. We set k to be 40 \% of the number of unseen classes. See text for more details.} \label{tb_KNN}
\vspace{2pt}
\small
\begin{tabular}{c|c|c|c}
\text{Distances for kNN using} & \textbf{AwA} & \textbf{CUB}  & \textbf{SUN} \\ 
& (k=4) & (k=20)  & (k=29) \\  \hline
Semantic representations & 57.5 & 68.9 & 75.2 \\ \hline
Predicted exemplars \hspace{2pt} & 67.5 & 80.0 & 82.1 \\ \hline
\end{tabular}
\vspace{5pt}
\end{table}

Besides the Pearson correlation coefficient used in Table 2 of the main text\footnote{We treat rows of each distance matrix as data points and compute the Pearson correlation coefficients between matrices.}, we provide another evidence that predicted exemplars better reflect visual similarities (as defined by real exemplars) than semantic representations. 
Let \%kNNoverlap(D) be the percentage of k-nearest neighbors (neighboring classes) using distances D that overlap with k-nearest neighbors using real exemplar distances. In Table~\ref{tb_KNN}, we report \%kNNoverlap (semantic representation distances) and \%kNNoverlap (predicted exemplar distances). We set k to be 40\% of the number of unseen classes, but we note that the trends are consistent for different k’s. Similar to the results in the main text, we observe clear improvement in all cases. 

\subsection{Larger visualization of the predicted exemplars}
\label{sSupplPredExem}

\begin{figure*}
\centering
\vspace{-20pt}
\includegraphics[width=.75\textwidth]{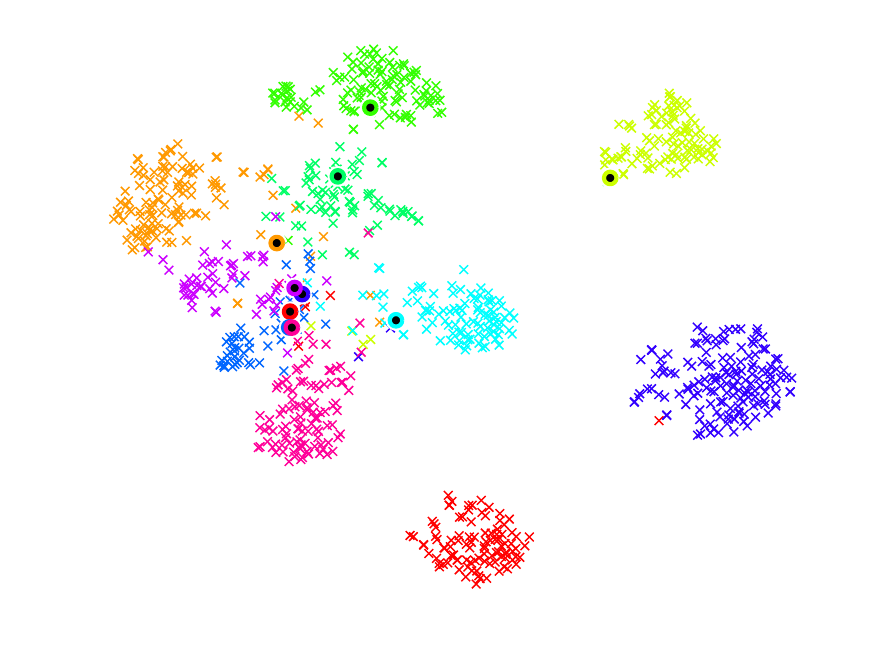}
\vspace{-25pt}
\caption{\small t-SNE \cite{tSNE} visualization of real images (crosses) and predicted visual exemplars (circles) for 10 \emph{unseen} classes on \textbf{AWA}. Different colors of symbols denote different unseen classes. Perfect predictions of visual features/exemplars would result in well-aligned crosses and circles of the same color. Best viewed in color.} \label{tSNE_AWA}
\vspace{-10pt}
\end{figure*}

\begin{figure*}
\centering
\includegraphics[width=.75\textwidth]{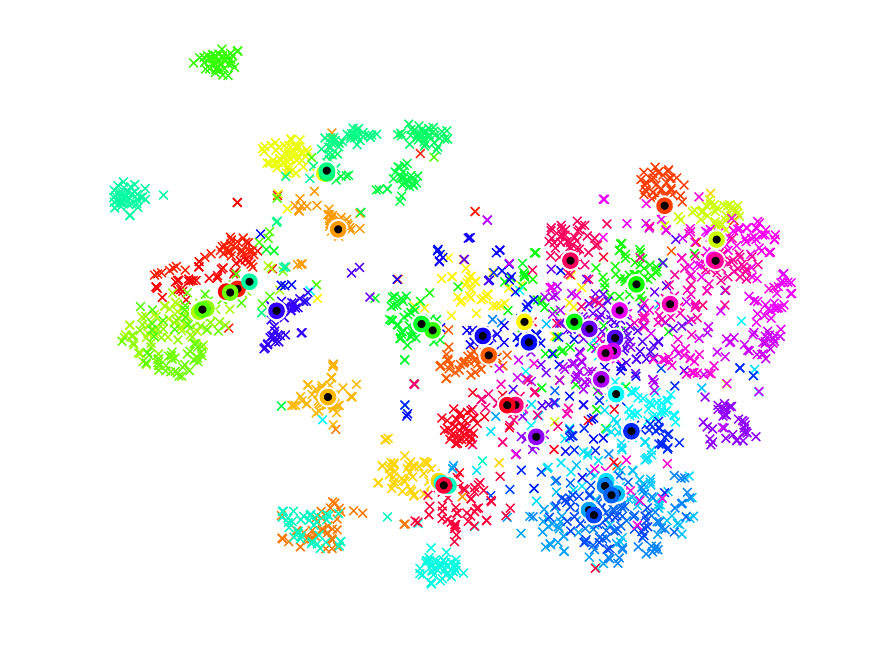}
\vspace{-25pt}
\caption{\small t-SNE \cite{tSNE} visualization of real images (crosses) and predicted visual exemplars (circles) for 50 \emph{unseen} classes on \textbf{CUB} (first split). Different colors of symbols denote different unseen classes. Perfect predictions of visual features/exemplars would result in well-aligned crosses and circles of the same color. Best viewed in color.} \label{tSNE_CUB}
\vspace{-10pt}
\end{figure*}

\begin{figure*}
\centering
\vspace{-20pt}
\includegraphics[width=.75\textwidth]{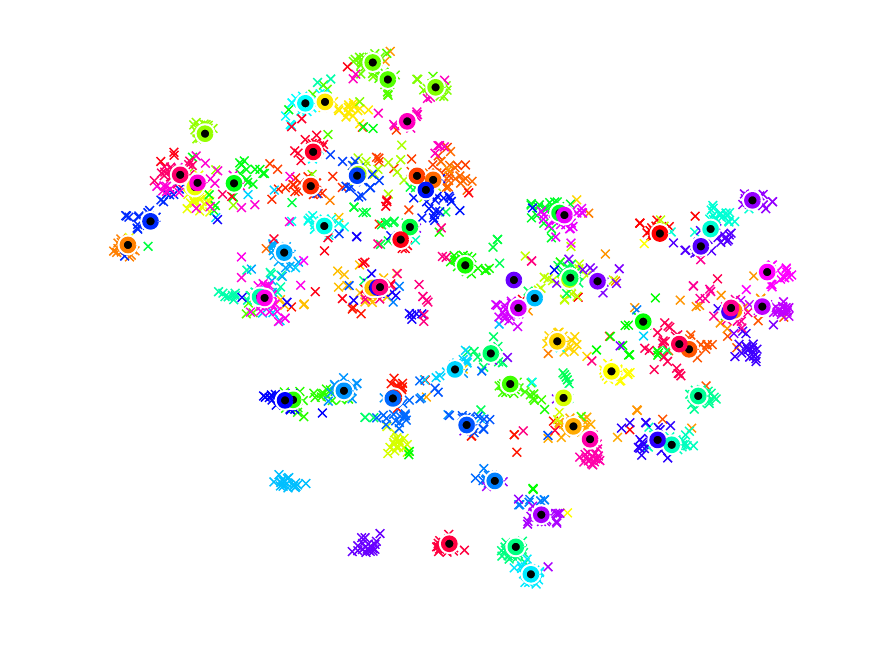}
\vspace{-20pt}
\caption{\small t-SNE \cite{tSNE} visualization of real images (crosses) and predicted visual exemplars (circles) for 72 \emph{unseen} classes on \textbf{SUN} (first split). Different colors of symbols denote different unseen classes. Perfect predictions of visual features/exemplars would result in well-aligned crosses and circles of the same color. Best viewed in color.} \label{tSNE_SUN}
\vspace{-10pt}
\end{figure*}

\begin{figure*}
\centering
\includegraphics[width=.75\textwidth]{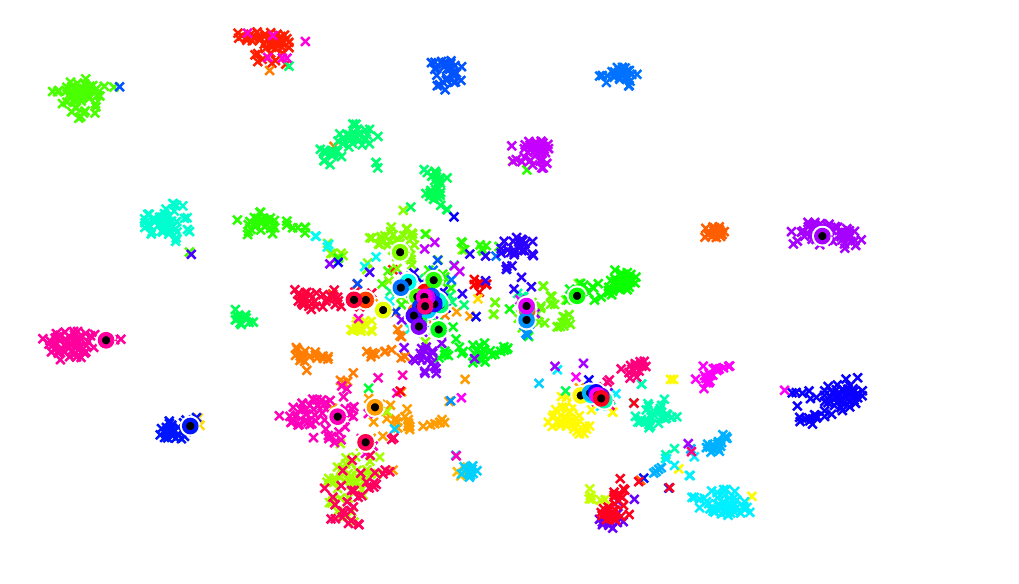}
\vspace{-10pt}
\caption{\small t-SNE \cite{tSNE} visualization of real images (crosses) and predicted visual exemplars (circles) for 48 randomly sampled \emph{unseen} classes from \emph{2-hop} on \textbf{ImageNet}. Different colors of symbols denote different unseen classes. Perfect predictions of visual features/exemplars would result in well-aligned crosses and circles of the same color. Best viewed in color.}
\label{tSNE_IMN}
\vspace{10pt}
\end{figure*}

We provide the t-SNE visualization~\cite{tSNE} of the predicted visual exemplars of the \emph{unseen} classes for \textbf{AwA}, \textbf{CUB}, \textbf{SUN}, and \textbf{ImageNet} in Fig.~\ref{tSNE_AWA}, \ref{tSNE_CUB}, \ref{tSNE_SUN}, and \ref{tSNE_IMN}, respectively --- each class is designated a color, with its corresponding real images/predicted exemplar marked with crosses/circle. 
Note that these figures are larger-size versions of Fig.~2 of the main text. For many of the unseen classes, the predicted exemplars are well aligned with their corresponding real images, explaining the superior performance of applying them for ZSL even though a simple nearest neighbor classification is used.

Note that it is the \emph{relative} distance that is important. Even when the predicted exemplars are not well aligned with their corresponding images, they are in many cases closer to those images than the predicted exemplars of other classes are. For example, on \textbf{AwA}, we would be able to predict test images from ``orange" class correctly as the closest exemplar is orange (but the images and the exemplar are not exactly aligned). 

\section{Expanded zero-shot learning results}
\label{sSupplZSLexp}

\subsection{\textbf{Expanded main results on small datasets}}

Table~\ref{tbMain_more} expands Table~3 of the main text to include additional baselines.
First, we include results of additional baselines \cite{AkataRWLS15,Bernardino15,MensinkGS14} reported in \cite{ChangpinyoCGS16}. 
Second, we report results of very recently proposed methods that use the more optimistic metric \emph{per-sample} accuracy as well as different types of deep visual features. 

\emph{Per-sample} accuracy is computed by averaging over accuracy of each sample. This is different from \emph{per-class} accuracy that is computed by averaging over accuracy of each unseen class. It is likely that \emph{per-sample} accuracy is the more optimistic metric of the two, as \cite{XianAS17} reports that they are unable to reproduce results of \mt{SSE} \cite{ZhangS15}, which uses \emph{per-sample} accuracy, with \emph{per-class} accuracy.

We also note that visual features can affect the performance greatly. For example, VGG features \cite{Simonyan15c} of \textbf{AwA} used in \cite{ZhangS15,ZhangS16,WangC16} are likely more discriminative than GoogLeNet features. In particular, \mt{BiDiLEL} \cite{WangC16} reports results on both features with VGG outperforming GoogLeNet by an absolute 5.8\%. This could explain strong results on \textbf{AwA} reported in \cite{ZhangS15,ZhangS16,WangC16}. It would also be interesting to investigate how GoogLetNet V2 \cite{IoffeS15} (in additional to \emph{per-sample} evaluation metric) used by \mt{DEm} \cite{ZhangXG16} contributes to their superior performance on \textbf{AwA}.

Finally, despite the variations in experimental settings, our method still outperforms all baselines on \textbf{CUB}.

\begin{table*}
\centering
\caption{\small Expanded comparison (cf. Table 3 of the main text) to existing ZSL approaches in the multi-way classification accuracies (in \%) on \textbf{AwA}, \textbf{CUB}, and \textbf{SUN}. For each dataset, we mark the best in red and the second best in blue. We include results of recent ZSL methods with other types of deep features (VGG by \cite{Simonyan15c} and GoogLeNet V2 by \cite{IoffeS15}) and/or different evaluation metrics. See text for details on how to interpret these results.} \label{tbMain_more}
\small
\begin{tabular}{c|c|c|c|c|c}
\text{Approach}	& Visual & Evaluation & \textbf{AwA} & \textbf{CUB} & \textbf{SUN} \\ 
	& features & metric & & 	& 	\\ \hline
\mtc{SSE} \cite{ZhangS15} & VGG	& per-sample & 76.3 & 30.4$^\S$ & - \\
\mtc{JLSE} \cite{ZhangS16} & VGG	& per-sample & {\color{blue}\textbf{80.5}} & 42.1$^\S$ & - \\
\mtc{BiDiLEL} \cite{WangC16} & VGG & per-sample & 79.1 & 47.6$^\S$ & - \\
\mtc{DEm} \cite{ZhangXG16}  & GoogLeNet V2 & per-sample & {\color{red}\textbf{86.7}} & 58.3$^\S$ & - \\  \hline
\mtc{SJE} \cite{AkataRWLS15} & GoogLeNet & per-class &  66.3 & 46.5 & 56.1\\
\mtc{ESZSL} \cite{Bernardino15} & GoogLeNet & per-class & 64.5 & 34.5& 18.7\\
\mtc{COSTA} \cite{MensinkGS14} & GoogLeNet & per-class & 61.8 & 40.8 & 47.9 \\
\mtc{ConSE}$^\dagger$ \cite{NorouziMBSSFCD14} & GoogLeNet & per-class & 63.3 & 36.2 & 51.9 \\
\mtc{BiDiLEL} \cite{WangC16} & GoogLeNet & per-class & 72.4 & 49.7$^\S$ & - \\
\mtc{LatEm}$^\ddagger$ \cite{XianASNHS16} & GoogLeNet & per-class & 72.1 & 48.0 & 64.5 \\
\mtc{SynC$^\textrm{o-vs-o}$} \cite{ChangpinyoCGS16} & GoogLeNet & per-class & 69.7 & 53.4 & 62.8 \\
\mtc{SynC$^\textrm{cs}$} \cite{ChangpinyoCGS16} & GoogLeNet & per-class & 68.4 	& 51.6 	& 52.9 \\
\mtc{SynC$^\textrm{struct}$} \cite{ChangpinyoCGS16}\hspace{2pt} & GoogLeNet & per-class & 72.9 	& 54.5 	& 62.7 \\ \hline
\mtc{EXEM (ConSE)} & GoogLeNet & per-class & 70.5 & 46.2  & 60.0 \\
\mtc{EXEM (LatEm)}$^\ddagger$ & GoogLeNet & per-class & 72.9 & 56.2  & {\color{blue}\textbf{67.4}} \\
\mtc{EXEM (SynC$^\textrm{o-vs-o}$)} & GoogLeNet & per-class & 73.8 & 56.2  & 66.5 \\
\mtc{EXEM (SynC$^\textrm{struct}$)} & GoogLeNet & per-class & 77.2 & {\color{red}\textbf{59.8}} & 66.1 \\
\mtc{EXEM (1NN)} \hspace{2pt} & GoogLeNet & per-class & 76.2	& 56.3 	& {\color{red}\textbf{69.6}} \\
\mtc{EXEM (1NNs)}\hspace{2pt} & GoogLeNet & per-class & 76.5 	& {\color{blue}\textbf{58.5}}	& {\color{blue}\textbf{67.3}} \\ \hline
\end{tabular}
\begin{flushleft}
$^\S$: on a particular split of seen/unseen classes. \hspace{5pt} $^\dagger$: reported in~\cite{ChangpinyoCGS16}. \hspace{5pt} $^\ddagger$: based on the code of~\cite{XianASNHS16}, averaged over 5 different initializations. 
\end{flushleft}
\vspace{-10pt}
\end{table*}

\subsection{Expanded \textbf{ImageNet} results}

Table~\ref{tbImagenet_more} expands the results of Table~4 in the main text to include other previously \emph{published} results that use AlexNet features \cite{KrizhevskySH12} and evaluate on all unseen classes. In all cases, our method outperforms the baseline approaches. 

\begin{table*}
\centering
\small 
\caption{\small Expanded comparison (cf. Table 4 of the main text) to existing ZSL approaches on \textbf{ImageNet} using \textbf{word vectors} of the class names as semantic representations. For both types of metrics (in \%), the higher the better. The best is in red. AlexNet is by \cite{KrizhevskySH12}. The number of actual unseen classes are given in parentheses. $^\dagger$: reported in~\cite{ChangpinyoCGS16}.}
\label{tbImagenet_more}
\small
\begin{tabular}{c|c|c|ccccc|cccc}
\text{Test data} & \text{Approach} & \text{Visual} & \multicolumn{5}{|c|}{Flat Hit@K} & \multicolumn{4}{|c}{Hierarchical precision@K}\\ \cline{4-12}
& K= & \text{features} & \text{1} & \text{2} & \text{5} & \text{10} & \text{20} & \text{2} & \text{5} & \text{10} & \text{20} \\ \hline
\emph{2-hop}
& \mtc{DeViSE}~\cite{FromeCSBDRM13} & AlexNet & {6.0} & {10.1} & {18.1} & {26.4} & {36.4} & {15.2} & {19.2} & {21.7} & {23.3} \\
(1,549)
& \mtc{ConSE}~\cite{NorouziMBSSFCD14} & AlexNet  & {9.4} & {15.1} & {24.7} & {32.7} & {41.8} & {21.4} & {24.7} & {26.9} & {28.4} \\  \hline
& \mtc{ConSE}$^\dagger$~\cite{NorouziMBSSFCD14} & GoogLeNet & 8.3 & 12.9 & 21.8 & 30.9 & 41.7 & 21.5 & 23.8 & 27.5 & 31.3 \\ 
& \mtc{SynC}$^\textrm{o-vs-o}$ \cite{ChangpinyoCGS16} & GoogLeNet & 10.5 & 16.7 & 28.6 & 40.1 & 52.0 & 25.1 & 27.7 & 30.3 & 32.1 \\
\emph{2-hop}
& \mtc{SynC}$^\textrm{struct}$ \cite{ChangpinyoCGS16} & GoogLeNet & 9.8 & 15.3 & 25.8 & 35.8 & 46.5 & 23.8 & 25.8 & 28.2 & 29.6 \\ \cline{2-12}
(1,509)
& \mtc{EXEM (SynC$^\textrm{o-vs-o}$)} & GoogLeNet & 11.8 & 18.9 & 31.8 & 43.2 & 54.8 & 25.6 & 28.1 & 30.2 & 31.6 \\
& \mtc{EXEM (1NN)} & GoogLeNet	& 11.7 & 18.3 & 30.9 & 42.7 & 54.8 & 25.9 & 28.5 & {\color{red}\textbf{31.2}} &{\color{red}\textbf{33.3}} \\
& \mtc{EXEM (1NNs)} & GoogLeNet & {\color{red}\textbf{12.5}} & {\color{red}\textbf{19.5}} & {\color{red}\textbf{32.3}} & {\color{red}\textbf{43.7}} & {\color{red}\textbf{55.2}} & {\color{red}\textbf{26.9}} & {\color{red}\textbf{29.1}} & 31.1 & 32.0 \\
\hline \hline 
\emph{3-hop}
& \mtc{DeViSE}~\cite{FromeCSBDRM13}& AlexNet  & {1.7} & {2.9} & {5.3} & {8.2} & {12.5} & {3.7} & {19.1} & {21.4} & {23.6} \\
(7,860)
& \mtc{ConSE}~\cite{NorouziMBSSFCD14}& AlexNet  & {2.7} & {4.4} & {7.8} & {11.5} & {16.1} & {5.3} & {20.2} & {22.4} & {24.7} \\ \hline 
& \mtc{ConSE}$^\dagger$~\cite{NorouziMBSSFCD14} & GoogLeNet & 2.6 & 4.1 & 7.3 & 11.1 & 16.4 & 6.7 & 21.4 & 23.8 & 26.3 \\ 
& \mtc{SynC}$^\textrm{o-vs-o}$ \cite{ChangpinyoCGS16} & GoogLeNet & 2.9 & 4.9 & 9.2 & 14.2 & 20.9 & 7.4 & 23.7 & 26.4 & 28.6 \\
\emph{3-hop}
& \mtc{SynC}$^\textrm{struct}$ \cite{ChangpinyoCGS16} & GoogLeNet & 2.9 & 4.7 & 8.7 & 13.0 & 18.6 & 8.0 & 22.8 & 25.0 & 26.7 \\ \cline{2-12}
(7,678)
& \mtc{EXEM (SynC$^\textrm{o-vs-o}$)} & GoogLeNet  & 3.4 & 5.6 & 10.3 & 15.7 & 22.8 & 7.5 & 24.7 & 27.3 & 29.5 \\ 
& \mtc{EXEM (1NN)} & GoogLeNet & 3.4 & 5.7 & 10.3 & 15.6 & 22.7 & 8.1 & {\color{red}\textbf{25.3}} & {\color{red}\textbf{27.8}} & {\color{red}\textbf{30.1}}	 \\
& \mtc{EXEM (1NNs)} & GoogLeNet & {\color{red}\textbf{3.6}} & {\color{red}\textbf{5.9}} & {\color{red}\textbf{10.7}} & {\color{red}\textbf{16.1}} & {\color{red}\textbf{23.1}} & {\color{red}\textbf{8.2}} & 25.2 & 27.7 & 29.9 \\
\hline \hline 
\emph{All}
& \mtc{DeViSE}~\cite{FromeCSBDRM13} & AlexNet  & {0.8} & {1.4} & {2.5} & {3.9} & {6.0} &{1.7} & {7.2} & {8.5} & {9.6} \\
(20,842)
& \mtc{ConSE}~\cite{NorouziMBSSFCD14} & AlexNet  & {1.4} & {2.2} & {3.9} & {5.8} & {8.3} & {2.5} & {7.8} & {9.2} & {10.4} \\ \hline
& \mtc{ConSE}$^\dagger$~\cite{NorouziMBSSFCD14} & GoogLeNet & 1.3 & 2.1 & 3.8 & 5.8 & 8.7 & 3.2 & 9.2 & 10.7 & 12.0 \\ 
& \mtc{SynC$^\textrm{o-vs-o}$} \cite{ChangpinyoCGS16} & GoogLeNet & 1.4 & 2.4 & 4.5 & 7.1 & 10.9 & 3.1 & 9.0 & 10.9 & 12.5 \\
\emph{All}
& \mtc{SynC$^\textrm{struct}$} \cite{ChangpinyoCGS16} & GoogLeNet & 1.5 & 2.4 & 4.4 & 6.7 & 10.0 & 3.6 & 9.6 & 11.0 & 12.2 \\ \cline{2-12}
(20,345)
& \mtc{EXEM (SynC$^\textrm{o-vs-o}$)} & GoogLeNet & 1.6 & 2.7 & 5.0 & 7.8 & 11.8 & 3.2 & 9.3 & 11.0 & 12.5 \\ 
& \mtc{EXEM (1NN)} & GoogLeNet & 1.7 & 2.8 & 5.2 & 8.1 & 12.1 & {\color{red}\textbf{3.7}} & {\color{red}\textbf{10.4}} & {\color{red}\textbf{12.1}} & {\color{red}\textbf{13.5}} \\
& \mtc{EXEM (1NNs)} & GoogLeNet  & {\color{red}\textbf{1.8}} & {\color{red}\textbf{2.9}} & {\color{red}\textbf{5.3}} & {\color{red}\textbf{8.2}} & {\color{red}\textbf{12.2}} & 3.6 & 10.2 & 11.8 & 13.2\\
\hline
\end{tabular}
\vspace{-10pt}
\end{table*}

\subsection{Additional ZSL results with word vectors as semantic representations}
In Table~\ref{tb_word2vec}, we show that we can improve the quality of word vectors on \textbf{AwA} as well. We use the 1,000-dimensional word vectors in~\cite{FuHXG15} and follow the same evaluation protocol as before. For other specific details, please refer to \cite{ChangpinyoCGS16}.

\begin{table}
\centering
\caption{\small ZSL results in the per-class multi-way classification accuracies (in \%) on \textbf{AwA} using word vectors as semantic representations. We use the 1,000-dimensional word vectors in~\cite{FuHXG15}. All approaches use GoogLeNet as the visual features.} \label{tb_word2vec}
\small
\begin{tabular}{c|c}
\text{Approach} & \textbf{AwA} \\ \hline
\mtc{SynC$^\textrm{o-vs-o}$} \cite{ChangpinyoCGS16} & 57.5 \\ \hline
\mtc{EXEM (SynC$^\textrm{o-vs-o}$)} & 61.7 \\
\mtc{EXEM (1NN)} \hspace{2pt} & 63.5 \\ \hline
\end{tabular}
\vspace{-3pt}
\end{table}

\subsection{Qualitative results}

\begin{figure*}
\centering
\includegraphics[width=.49\textwidth]{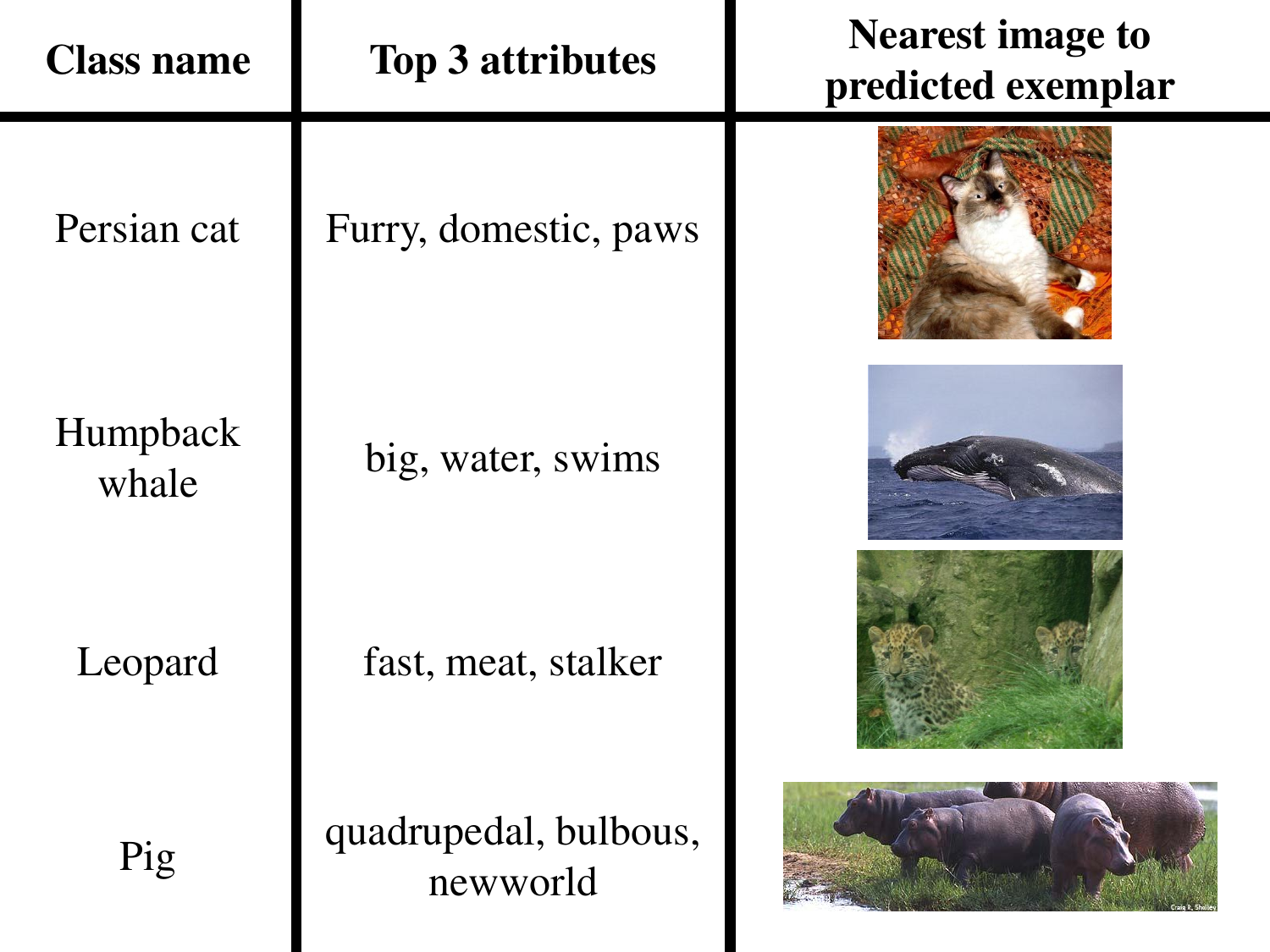}
\includegraphics[width=.49\textwidth]{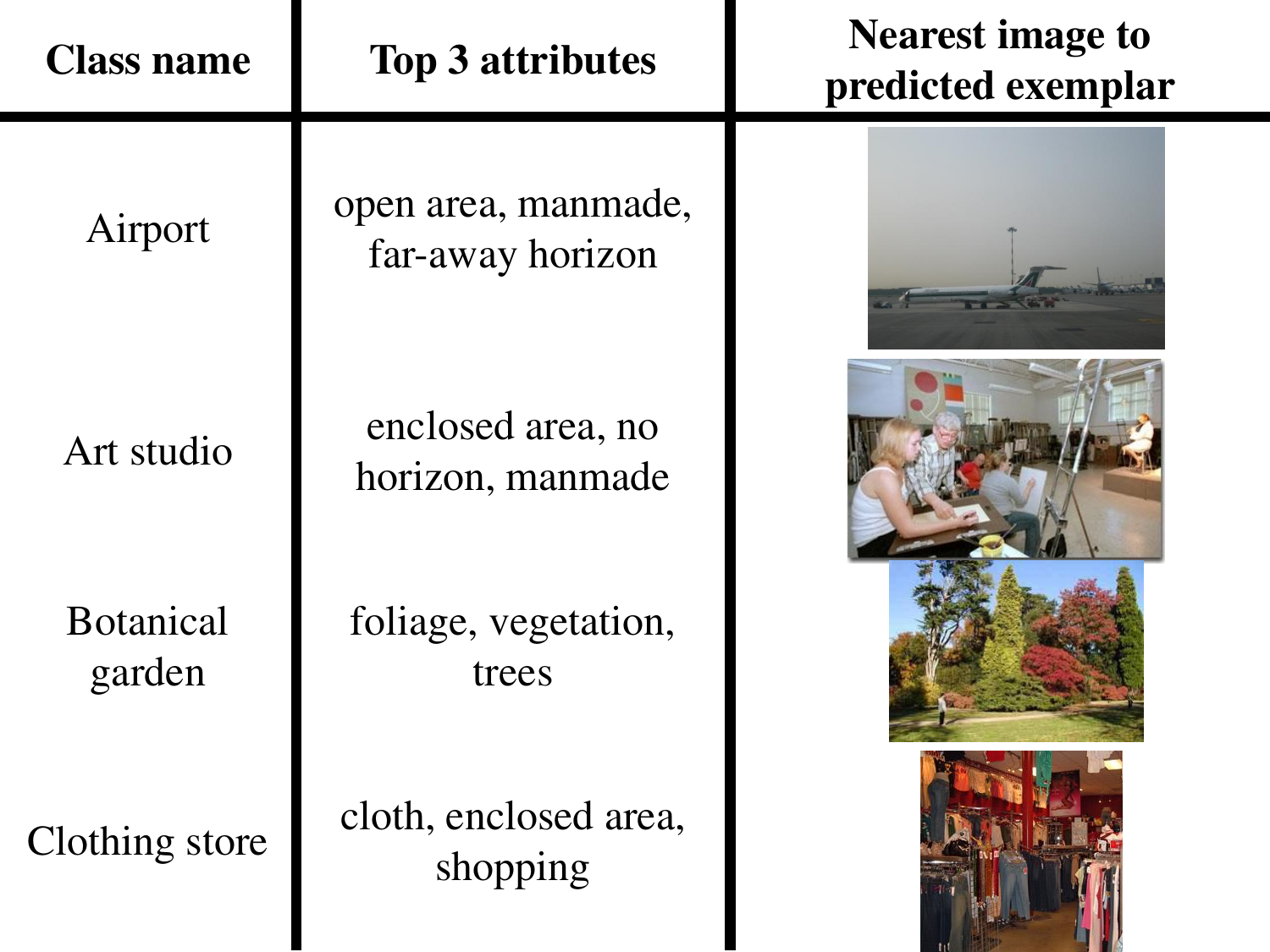}
\caption{\small Qualitative zero-shot learning results on \textbf{AwA} (left) and \textbf{SUN} (right). For each row, we provide a class name, three attributes with the highest strength, and the nearest image to the predicted exemplar (projected back to the original visual feature space).}
\label{fQual}
\vspace{-10pt}
\end{figure*}

Finally, we provide qualitative results on the zero-shot learning task on \textbf{AwA} and \textbf{SUN} in Fig.~\ref{fQual}. 
For each row, we provide a class name, three attributes with the highest strength, and the nearest image to the predicted exemplar (projected back to the original visual feature space).
We stress that each class that we show here is an \emph{unseen} class, and the images are from unseen classes as well.
Generally, the results are reasonable; class names, attributes, and images generally correspond well. Even when the image is from the wrong class, the appearance of the nearest image is reasonable. For example, we predict a hippopotamus exemplar from the pig attributes, but the image does not look too far from pigs. This could also be due to the fact that many of these attributes are not \emph{visual} and thus our regressors are prone to learning the wrong thing \cite{Jayaraman14}.   

\section{Generalized zero-shot learning results}
\label{sSupplGZSLexp}

\begin{table}
\centering
\caption{\small Generalized ZSL results in Area Under Seen-Unseen accuracy Curve (AUSUC)~\cite{ChaoCGS16} on \textbf{AwA}, \textbf{CUB}, and \textbf{SUN}. For each dataset, we mark the best in red and the second best in blue. All approaches use GoogLeNet as the visual features and calibrated stacking~\cite{ChaoCGS16} to combine the scores for seen and unseen classes.} \label{tb_GZSL}
\small
\begin{tabular}{c|c|c|c}
\text{Approach} & \textbf{AwA} & \textbf{CUB} & \textbf{SUN} \\  \hline
\mtc{DAP}$^\dagger$ \cite{LampertNH14}  & 0.366 & 0.194 & 0.096 \\
\mtc{IAP}$^\dagger$ \cite{LampertNH14}  & 0.394 & 0.199 & 0.145 \\
\mtc{ConSE}$^\dagger$ \cite{NorouziMBSSFCD14}  & 0.428 & 0.212 & 0.200 \\
\mtc{ESZSL}$^\dagger$ \cite{Bernardino15}  & 0.449 & 0.243 & 0.026 \\
\mtc{SynC$^\textrm{o-vs-o}$}$^\dagger$ \cite{ChangpinyoCGS16} & 0.568 & 0.336 & 0.242 \\
\mtc{SynC$^\textrm{struct}$}$^\dagger$ \cite{ChangpinyoCGS16}\hspace{2pt} & 0.583 	& 0.356 	& 0.260 \\ \hline\
\mtc{EXEM (SynC$^\textrm{o-vs-o}$)}  & 0.553 & 0.365  & 0.265 \\
\mtc{EXEM (SynC$^\textrm{struct}$)}  & {\color{red}\textbf{0.587}} & {\color{red}\textbf{0.397}} & {\color{red}\textbf{0.288}} \\
\mtc{EXEM (1NN)} \hspace{2pt} & 0.570	& 0.318 	& 0.284 \\
\mtc{EXEM (1NNs)}\hspace{2pt}  & {\color{blue}\textbf{0.584}} 	& {\color{blue}\textbf{0.373}}	& {\color{blue}\textbf{0.287}} \\ \hline
\end{tabular}
\begin{flushleft}
$^\dagger$: results are reported in~\cite{ChaoCGS16}.
\end{flushleft}
\vspace{-10pt}
\end{table}

Conventional zero-shot learning setting unrealistically assumes that test data always come from the unseen classes.
Motivated by this, recent work proposes to evaluate zero-shot learning methods in the more practical setting called generalized zero-shot learning (GZSL). In GZSL, instances from both seen and unseen classes are present at test time, and the label space is the union of both types of classes. We refer the reader for more discussions regarding GZSL and related settings in \cite{ChaoCGS16,XianAS17}.

We evaluate our methods and baselines using the Area Under Seen-Unseen accuracy Curve (AUSUC) \cite{ChaoCGS16} and report the results in Table~\ref{tb_GZSL}. Following the same evaluation procedure as before, our approach again outperforms the baselines on all datasets.

Recently, Xian et al. \cite{XianAS17} proposes to unify the evaluation protocol in terms of image features, class semantic embeddings, data splits, and evaluation criteria for conventional and generalized zero-shot learning. In their protocol, GZSL is evaluated by the harmonic mean of seen and unseen classes' accuracies. Technically, AUSUC provides a more complete picture of zero-shot learning method's performance, but it is less simpler than the harmonic mean. As in Sect.~\ref{sSuppZSLBaselines}, further investigation under this newly proposed evaluation protocol (both in conventional and generalized zero-shot learning) is left for future work.

\section{Additional details on zero-shot to few-shot learning experiments}
\label{sSupplPeeked}

\subsection{Details on how to select peeked unseen classes}
\label{peekedunseen_select}
Denote by $B$ the number of peeked unseen classes whose labeled data will be revealed. In what follows, we provide detailed descriptions of how we select a subset of peeked unseen classes of size $B$.
\paragraph{Uniform random and heavy (light)-toward-seen random}
As mentioned in Sect.~3.1 of the main text, there are different subsets of unseen classes on \textbf{ImageNet} according to the WordNet hierarchy. Each subset contains unseen classes with a certain range of tree-hop distance from the 1K seen classes. The smaller the distance is, the higher the semantic similarity between unseen classes and seen classes.
Here, we consider the following three \emph{disjoint} subsets:
\begin{itemize}[noitemsep]
\item \emph{2-hop}: 1,509 (out of 1,549) unseen classes
that are within 2 tree-hop distance from the 1K seen classes.
\item \emph{Pure 3-hop}: 6,169 (out of 6,311) unseen classes that are with exactly 3 tree-hop distance from the 1K seen classes.
\item \emph{Rest}: 12,667 (out of 12,982) unseen classes that are with more than 3 tree-hop distance from the 1K seen classes.
\end{itemize} 

Note that \emph{3-hop} defined in Sect.~3.1 of the main text is exactly $\emph{2-hop} \cup \emph{Pure 3-hop}$, and \emph{All} is $\emph{2-hop} \cup \emph{Pure 3-hop} \cup \emph{Rest}$.

For uniform random, we pick from \emph{2-hop}/\emph{Pure 3-hop}/ \emph{Rest} the number of peeked unseen classes proportional to their set size (i.e., 1,509/6,169/12,667). That is, we do not bias the selected classes towards any subset. For heavy-toward-seen random, we pick from \emph{2-hop}/\emph{Pure 3-hop}/\emph{Rest} the number of peeked unseen classes proportional to (16$\times$1,509)/(4$\times$6,169)/(1$\times$12,667). For light-toward-seen random, we pick from \emph{2-hop}/\emph{Pure 3-hop}/\emph{Rest} the number of peeked unseen classes proportional to (1$\times$1,509)/(4$\times$6,169)/(16$\times$12,667). Given the number of peeked unseen classes for each subset, we then perform uniform sampling (without replacement) within each subset to select the peeked unseen classes. If the number of peeked unseen classes to select from a subset exceeds the number of classes of that subset, we split the exceeding budget to other subsets following the proportion.

\paragraph{DPP}
Given a ground set of $\cN$ items (e.g., classes) and the corresponding $\cN$-by-$\cN$ kernel matrix $\mat{L}$ that encodes the pair-wise item similarity, a DPP~\cite{kulesza2012determinantal} defines the probability of any subset sampled from the ground set. The probability of a specific subset is proportional to the determinant of the principal minor of $\mat{L}$ indexed by the subset. A diverse subset is thus with a higher probability to be sampled.

For zero-shot to few-shot learning experiments, we construct $\mat{L}$ with the RBF kernel computed on semantic representations (e.g, word vectors) of all the seen and unseen classes (i.e., $\cS+\cU$ classes). We then compute the $\cU$-by-$\cU$ kernel matrix $\mat{L}_U$ conditional on that all the $\cS$ seen classes are already included in the subset. Please refer to~\cite{kulesza2012determinantal} for details on conditioning in DPPs. With $\mat{L}_U$, we would like to select additional $B$ classes that are diverse from each other and from the seen classes to be the peeked unseen classes.

Since finding the most diverse subset (either fixed-size or not) is an NP-hard problem~\cite{KuleszaT11, kulesza2012determinantal}, we apply a simple greedy algorithm to sequentially select classes. Denote $Q_t$ as the set of peeked unseen classes with size $t$ and $U_t$ as the remaining unseen classes, we enumerate all possible subset of size $t+1$ (i.e., $Q_t \cup \{c\in U_t\}$). We then include $c^*$ that leads to the largest probability into $Q_{t+1}$ (i.e., $Q_{t+1}=Q_t \cup \{c^*\}$ and $U_{t+1}=U_t - \{c^*\}$). We iteratively perform the update until $t = B$. 

\subsection{Additional results on zero-shot to few-shot learning results}

In this section, we analyze experimental results for \mt{EXEM (1NN)} in detail. We refer the reader to the setup described in Sect.~3.3.3 of the main text, as well as additional setup below.

\subsubsection{Additional setup}
We will consider several fine-grained evaluation metrics. We denote by $\mathcal{A}^K_{\mathcal{X} \rightarrow \mathcal{Y}}$ the Flat Hit@K of classifying test instances from $\mathcal{X}$ to the label space of $\mathcal{Y}$. 
Since there are two types of test data, we will have two types of accuracy: the peeked unseen accuracy $\mathcal{A}^K_{\mathcal{P} \rightarrow \mathcal{U}}$ and the remaining unseen accuracy $\mathcal{A}^K_{\mathcal{R} \rightarrow \mathcal{U}}$, where $\mathcal{P}$ is the peeked unseen set, $\mathcal{R}$ is the remaining unseen set, and $\mathcal{U} = \mathcal{P} \cup \mathcal{R}$ is the unseen set. Then, the combined accuracy $\mathcal{A}^K_{\mathcal{U} \rightarrow \mathcal{U}} = w_{\mathcal{P}} \mathcal{A}^K_{\mathcal{P} \rightarrow \mathcal{U}} + w_{\mathcal{R}} \mathcal{A}^K_{\mathcal{R} \rightarrow \mathcal{U}}$, where $w_{\mathcal{P}}$ ($w_{\mathcal{R}}$) is the proportion of test instances in peeked unseen (remaining unseen) classes to the total number of test instances. Note that the combined accuracy is the one we use in the main text.

Note also that we follow the evaluation protocol for \textbf{ImageNet} in previous literature by using ``per-image" accuracy. We will also explore ``per-class" accuracy and show that we reach the same conclusion. See Sect~\ref{sSupplAb}.

\subsubsection{Full curves for \mt{EXEM (1NN)}}

\begin{figure}[t]
\centering
\includegraphics[width=0.45\textwidth]{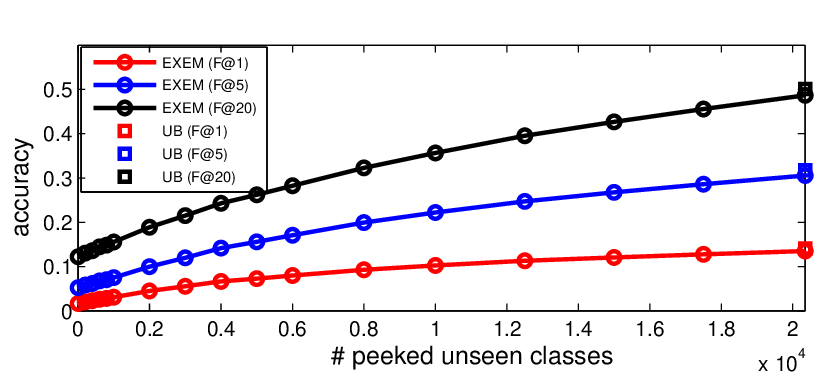}
\caption{\small Combined accuracy $\mathcal{A}^K_{\mathcal{U} \rightarrow \mathcal{U}}$ vs. the number of peeked unseen classes for \mtc{EXEM (1NN)}. The ``squares" correspond to \textbf{the upperbound (UB)} obtained by \mtc{EXEM (1NN)} on \emph{real} exemplars. F@K=1, 5, and 20.} \label{fAccEXEM}
\vspace{5pt}
\end{figure}

Fig.~\ref{fAccEXEM} shows $\mathcal{A}^K_{\mathcal{U} \rightarrow \mathcal{U}}$ when the number of peeked unseen classes keeps increasing. We observe this leads to improved overall accuracy, although the gain eventually is flat. We also show the upperbound: \mt{EXEM (1NN)} with \emph{real exemplars} instead of predicted ones for all the unseen classes. Though small, the gap to the upperbound could potentially be improved with a more accurate prediction method of visual exemplars, in comparison to SVR (Sect.~\ref{sSupplApproach}).

\subsubsection{Detailed analysis of the effect of labeled data from peeked unseen classes}

\begin{figure}[t]
\centering
\includegraphics[width=0.49\textwidth]{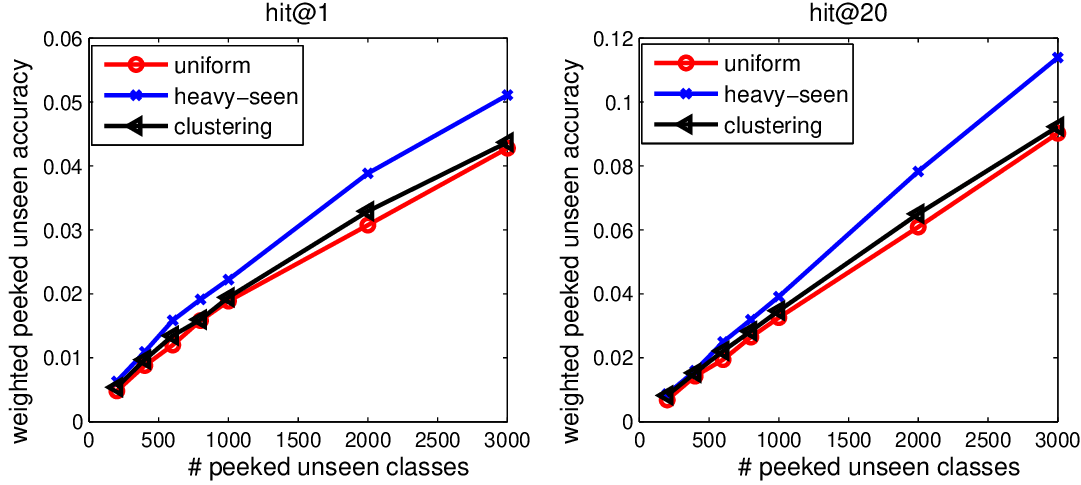}
\includegraphics[width=0.49\textwidth]{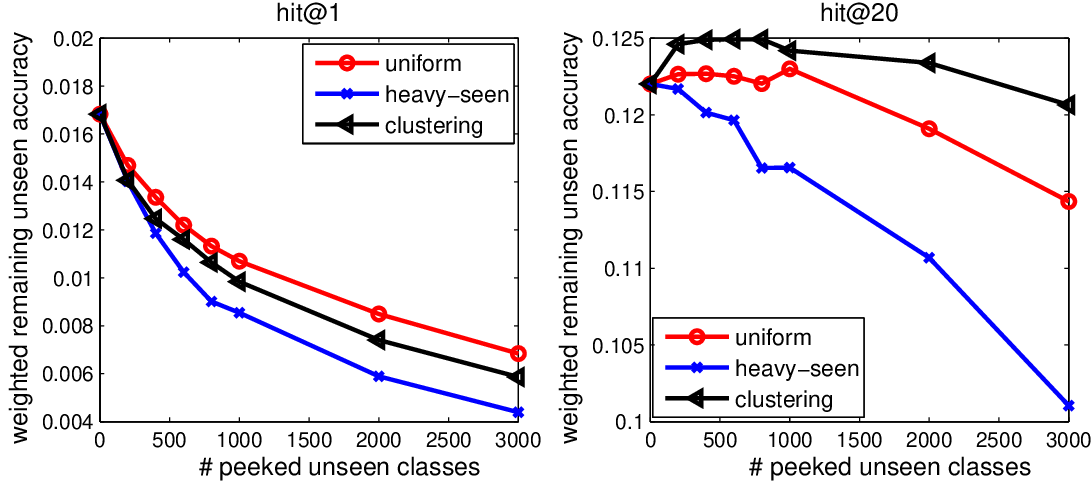}
\caption{\small (Top) Weighted \textbf{peeked unseen} accuracy $w_{\mathcal{P}} \mathcal{A}^K_{\mathcal{P} \rightarrow \mathcal{U}}$ vs. the number of peeked unseen classes. (Bottom) Weighted \textbf{remaining unseen} $w_{\mathcal{R}} \mathcal{A}^K_{\mathcal{R} \rightarrow \mathcal{U}}$ accuracy vs. the number of peeked unseen classes. The weight $w_{\mathcal{P}}$ ($w_{\mathcal{R}}$) is the number of test instances belonging to $\mathcal{P}$ ($\mathcal{R}$) divided by the total number of test instances. The evaluation metrics are F@1 (left) and F@20 (right).} \label{fPeekedUnseenAcc}
\end{figure}

\begin{figure}[t]
\centering
\includegraphics[width=0.49\textwidth]{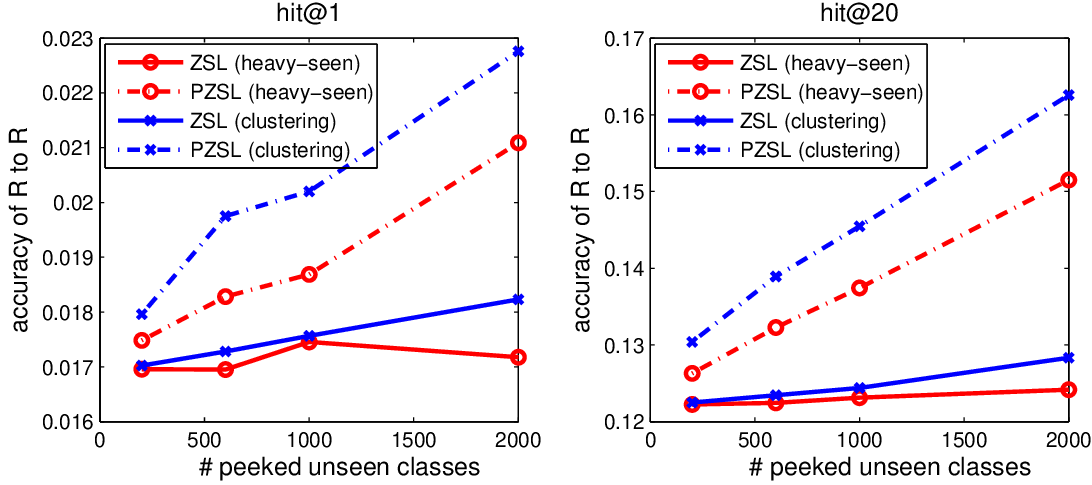}
\caption{\small Accuracy on test instances from the \textbf{remaining unseen classes when classifying into the label space of remaining unseen classes only} $\mathcal{A}^K_{\mathcal{R} \rightarrow \mathcal{R}}$ vs. the number of peeked unseen classes. ZSL trains on labeled data from the seen classes only while PZSL (ZSL with peeked unseen classes) trains on the the labeled data from both seen and peeked unseen classes. The evaluation metrics are F@1 (left) and F@20 (right).} \label{fRemainAccNobias}
\end{figure}

\begin{figure}[t]
\centering
\includegraphics[width=0.49\textwidth]{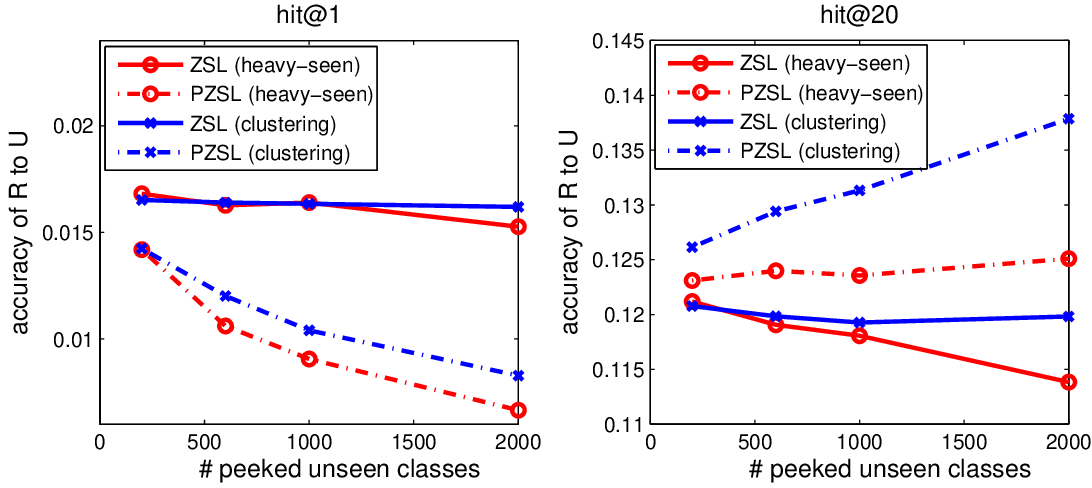}
\caption{\small Accuracy on test instances from the \textbf{remaining unseen classes when classifying into the label space of unseen classes} $\mathcal{A}^K_{\mathcal{R} \rightarrow \mathcal{U}}$ vs. the number of peeked unseen classes. ZSL trains on labeled data from the seen classes only while PZSL (ZSL with peeked unseen classes) trains on the the labeled data from both seen and peeked unseen classes. Note that these plots are the unweighted version of those at the bottom row of Fig.~\ref{fPeekedUnseenAcc}. The evaluation metrics are F@1 (left) and F@20 (right).} \label{fRemainAccWithbias}
\end{figure}

Fig.~\ref{fPeekedUnseenAcc} expands the results in Fig.~\ref{fAccEXEM} by providing the weighed peeked unseen accuracy $w_{\mathcal{P}} \mathcal{A}^K_{\mathcal{P} \rightarrow \mathcal{U}}$ and the weighted remaining unseen accuracy $w_{\mathcal{R}} \mathcal{A}^K_{\mathcal{R} \rightarrow \mathcal{U}}$. We note that, as the number of peeked unseen classes increases, $w_{\mathcal{P}}$ goes up while $w_{\mathcal{R}}$ goes down, roughly linearly in both cases. Thus, the curves go up for the top row and go down for the bottom row.

As we observe additional labeled data from more peeked unseen classes, the weighed peeked unseen accuracy improves roughly linearly as well. On the other hand, the weighed remaining unseen accuracy degrades very quickly for F@1 but slower for F@20. 
This suggests that the improvement we see (over ZSL performance) in Fig.~4 of the main text and Fig.~\ref{fAccEXEM} is contributed largely by the fact that peeked unseen classes benefit themselves. \emph{But how do peeked unseen classes exactly affect the remaining unseen classes?}

The above question is tricky to answer. There are two main factors that contribute to the performance on remaining unseen test instances.
The first factor is the confusion among remaining classes themselves, and the second one is the confusion with peeked unseen classes. We perform more analysis to understand the effect of each factor when classifying test instances from the remaining unseen set $\mathcal{R}$.

To remove the confusion with peeked unseen classes, we first restrict the label space to only the remaining unseen classes $\mathcal{R}$. In particular, we consider $\mathcal{A}^K_{\mathcal{R} \rightarrow \mathcal{R}}$ and compare the method in two settings: ZSL and PZSL (ZSL with peeked unseen classes). ZSL uses only the training data from seen classes while PZSL uses the training data from both seen and peeked unseen classes. In Fig.~\ref{fRemainAccNobias}, we see that adding labeled data from peeked unseen classes \textbf{\emph{does help}} by resolving confusion among remaining unseen classes \emph{themselves}, suggesting that peeked unseen classes inform other remaining unseen classes about visual information.

In Fig.~\ref{fRemainAccWithbias}, we add the confusion introduced by peeked unseen classes back by letting the label space consist of both $\mathcal{P}$ and $\mathcal{R}$. That is, we consider $\mathcal{A}^K_{\mathcal{R} \rightarrow \mathcal{U}}$.
For Flat Hit@1, the accuracy is hurt so badly that it goes down below ZSL baselines. However, for Flat Hit@20, the accuracy drops but still higher than ZSL baselines. 

Thus, to summarize, adding peeked unseen classes has two benefits: it improves the accuracies on the peeked unseen classes $\mathcal{P}$ (Fig.~\ref{fPeekedUnseenAcc} (Top)), as well as reduces the confusion among remaining unseen classes $\mathcal{R}$ themselves (Fig.~\ref{fRemainAccNobias}).  It biases the resulting classifiers towards the peeked unseen classes, hence causing confusion between $\mathcal{P}$ and $\mathcal{R}$ (Fig.~\ref{fRemainAccWithbias}). When the pros outweigh the cons, we observe overall improvement (Fig.~\ref{fAccEXEM}). 
Additionally, when we use less strict metrics, peeked unseen classes always help (Fig.~\ref{fRemainAccWithbias}).

\subsubsection{Results on additional metric, additional method, and additional rounds}
\label{sSupplAb}

We further provide experimental results on additional metric: per-class accuracy and on multiple values of K in Flat Hit @K (i.e., K $\in \{1,2,5,10,20\}$); additional method: \mt{EXEM (1NNs)}. We also provide results for \mt{EXEM (1NN)} averaged over multiple rounds using heavy-toward-seen random, light-toward-seen random, and uniform random to select peeked unseen classes to illustrate the stability of these methods.

\begin{figure*}
\centering
\includegraphics[width=0.95\textwidth]{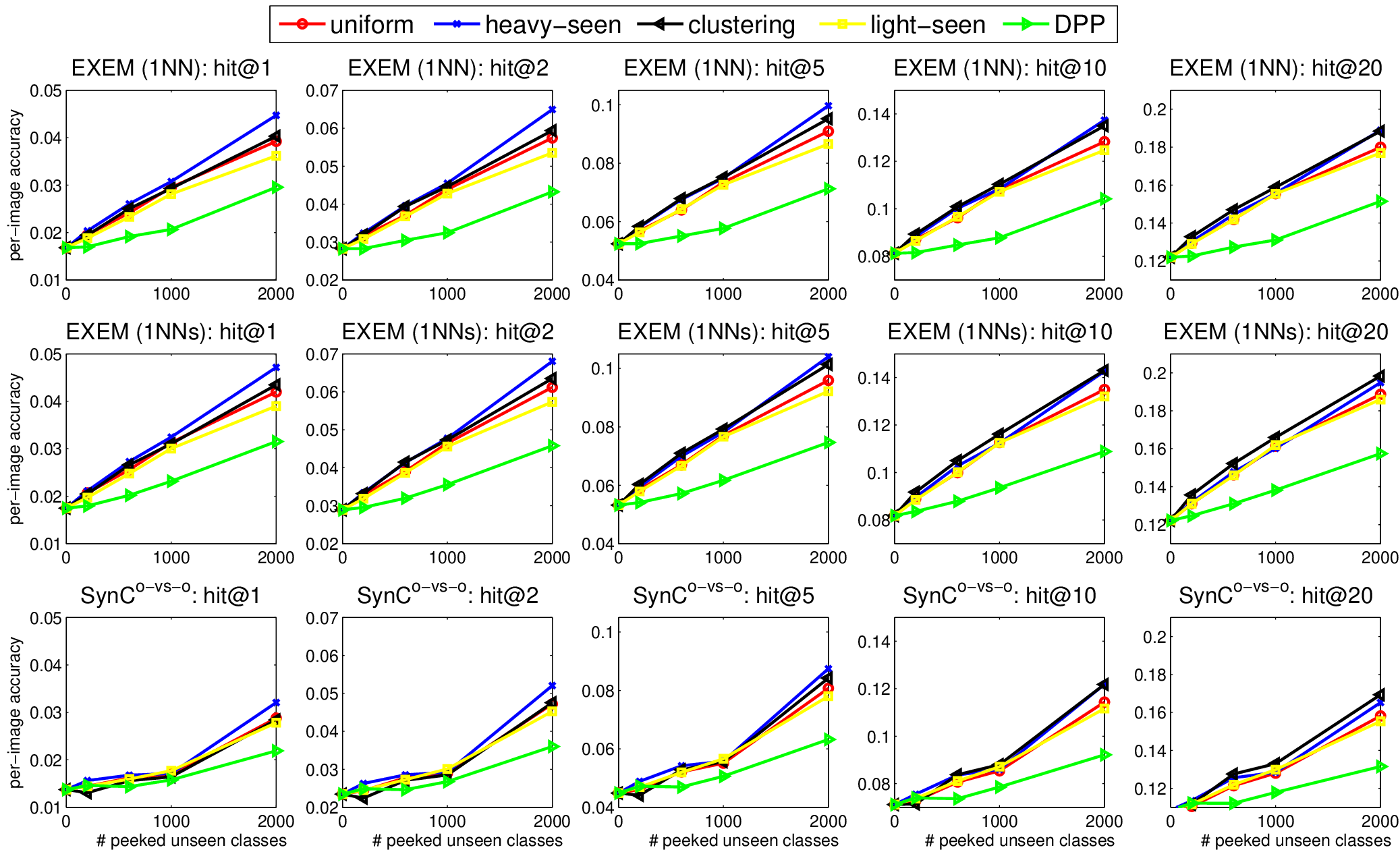}
\caption{\small Combined per-image accuracy vs. the number of peeked unseen classes for \mtc{EXEM (1NN)}, \mtc{EXEM (1NNs)}, and \mtc{SynC}. The evaluation metrics are, from left to right, Flat Hit@1 ,2 ,5, 10, and 20. Five subset selection approaches are considered.} \label{per_sample}
\vspace{-3pt}
\end{figure*}

\begin{figure*}
\centering
\includegraphics[width=0.95\textwidth]{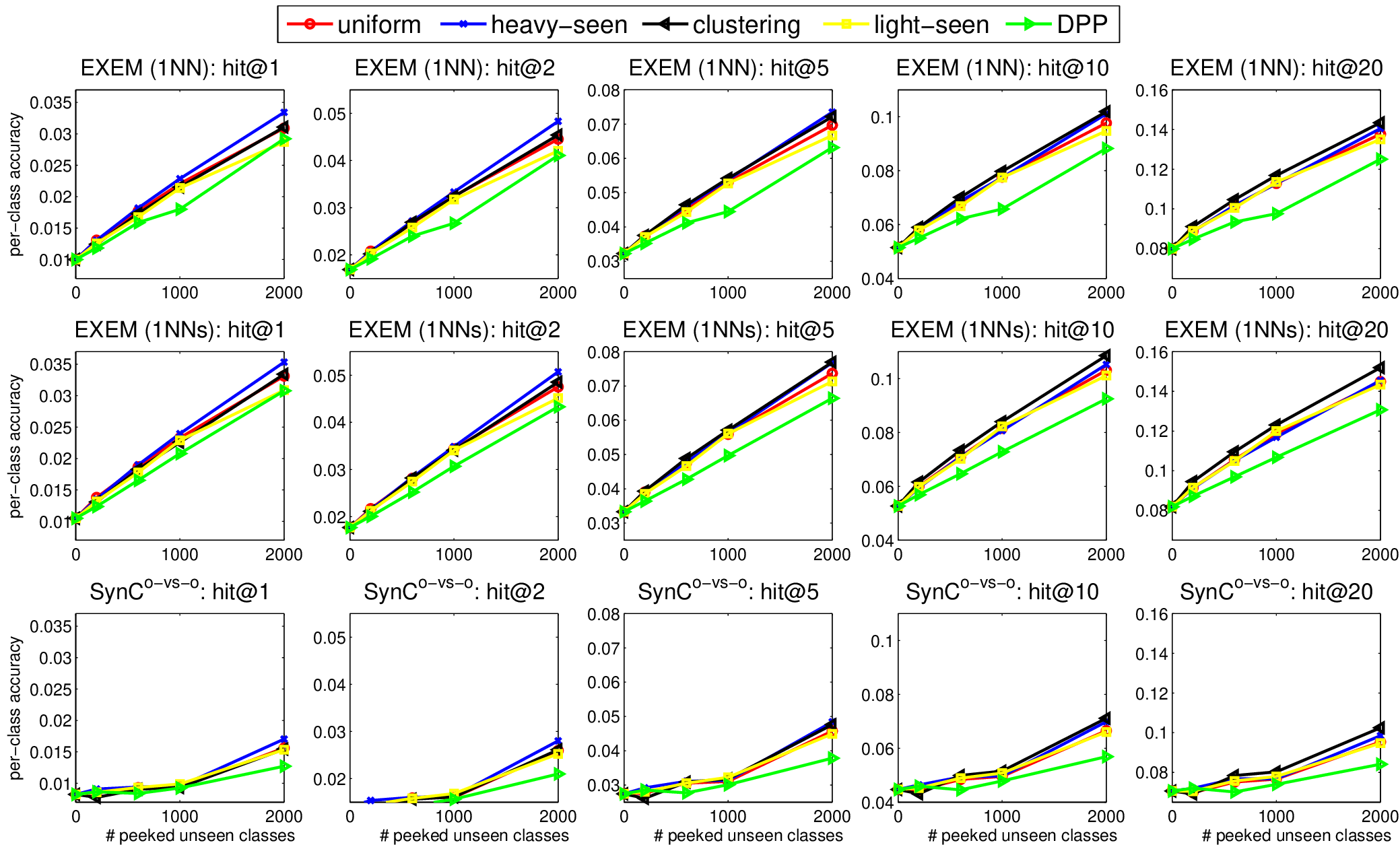}
\caption{\small Combined per-class accuracy vs. the number of peeked unseen classes for \mtc{EXEM (1NN)}, \mtc{EXEM (1NNs)}, and \mtc{SynC}. The evaluation metrics are, from left to right, Flat Hit@1, 2, 5, 10, and 20. Five subset selection approaches are considered.} \label{per_class}
\vspace{-3pt}
\end{figure*}

Fig. \ref{per_sample} and \ref{per_class} summarize the results for per-image and per-class accuracy, respectively. 
For both figures, each row corresponds to a ZSL method and each column corresponds to a specific value of K in Flat Hit@K.
In particular, from top to bottom, ZSL methods are \mt{EXEM (1NN)}, \mt{EXEM (1NNs)}, and \mt{SynC$^{\text{o-vs-o}}$}. From left to right, Flat Hit@K = 1, 2, 5, 10, and 20.

\paragraph{Where to peek?} No matter which type of accuracy is considered, we observe similar trends previously seen in the main text. \emph{heavy-toward-seen} is preferable for strict metrics (i.e., small K) while \emph{clustering} is preferable for flexible metrics (i.e., large K), for all zero-shot learning algorithms.

\paragraph{Comparison between ZSL methods}  No matter which type of accuracy is considered, we observe similar trends seen in the main text. \mt{EXEM (1NNs)}, under the same Flat Hit@K and subset selection method and number of peeked unseen classes, slightly outperforms \mt{EXEM (1NN)}. Both \mt{EXEM (1NN)} and \mt{EXEM (1NNs)} outperform \mt{SynC$^{\text{o-vs-o}}$}.

\paragraph{Per-class accuracy vs. Per-image accuracy} Per-class accuracy is generally lower than per-image accuracy. This can be attributed to two factors. First, the average number of instances per class in \emph{2-hop} is larger than that in \emph{Pure 3-hop} and \emph{Rest} (see Sect.~\ref{peekedunseen_select} for the definition)\footnote{On average, \emph{2-hop} has 696 instances/class, \emph{Pure 3-hop} has 584 instances/class, and \emph{Rest} has 452 nstances/class}. Second, the per-class accuracy in \emph{2-hop} is higher than that in \emph{Pure 3-hop} and \emph{Rest}\footnote{For example, the per-class accuracy of \mtt{EXEM (1NN)} in \emph{2-hop}/\emph{Pure 3-hop}/\emph{Rest} is 12.1/3.0/0.8 (\%) at Flat Hit@1 under 1,000 peeked unseen classes selected by heavy-toward-seen random.}. That is, when we compute the per-image accuracy, we emphasize the accuracy from \emph{2-hop}. The first factor indicates the long-tail phenomena in \textbf{ImageNet}, and the second factor indicates the nature of zero-shot learning --- unseen classes that are semantically more similar to the seen ones perform better than those that are less similar.

\begin{figure}
\centering
\includegraphics[width=0.42\textwidth]{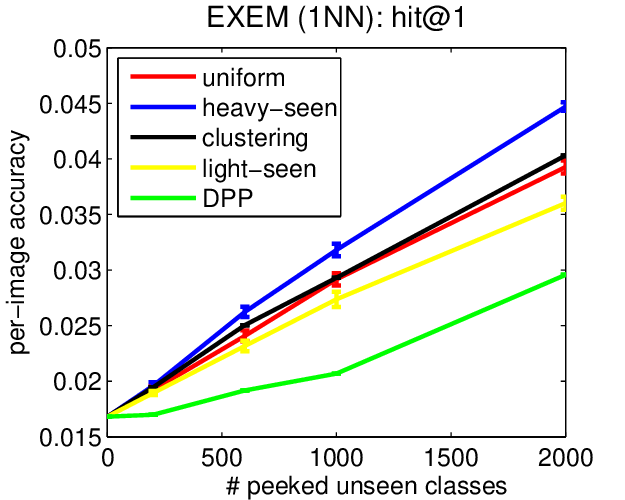}
\includegraphics[width=0.42\textwidth]{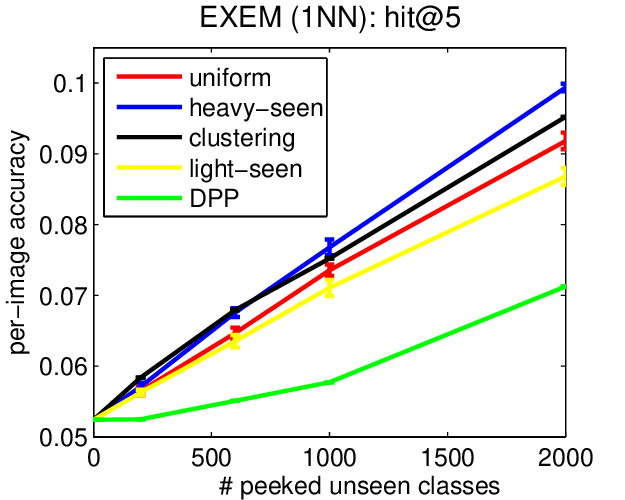}
\includegraphics[width=0.42\textwidth]{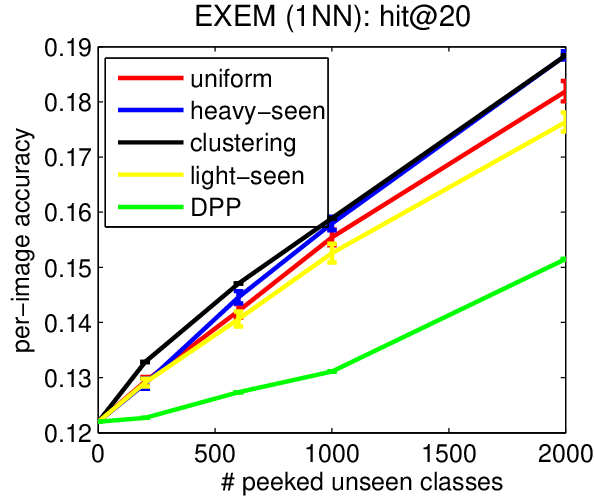}
\caption{\small Stability of random subset selection strategies. Combined per-image accuracy vs. the number of peeked unseen classes for \mtc{EXEM (1NN)}. Five subset selection approaches are compared. The results of heavy-toward-seen, light-toward-seen, and uniform random are \textbf{averaged over 10 random rounds} along with the error bars showing the standard deviation. The evaluation metrics are, from top to bottom, Flat Hit@1, 5, and 20.} \label{plot_stability}
\vspace{-5pt}
\end{figure}

\paragraph{Stability of peeked unseen class random selection}
For all experimental results on PZSL above and in the main text, we use a single round of randomness for heavy-toward-seen, light-toward-seen, and uniform random (see Sect.~\ref{peekedunseen_select} for details). That is, given budget $B$, we apply a fixed set of peeked unseen classes sampled according to a particular random strategy to all ZSL methods. To illustrate the stability of these random methods, we consider \mt{EXEM (1NN)} with 10 rounds of randomness and provide the mean and standard deviation of accuracy in Fig.~\ref{plot_stability}. 

The results follow the same trends as shown in Fig.~\ref{per_sample} and Fig.~4 of the main text; the standard deviation is small when compared to the gap among different random selection methods. This observation suggests that, for random subset selection, the distribution of peeked unseen classes is more important than specific peeked unseen classes being selected.

\paragraph{Real or predicted exemplars for peeked classes}

\begin{table}
\centering
\caption{\small Comparison on using predicted and real exemplars for the peeked classes for few-shot learning (in \%). \mtc{EXEM (1NN)} with heavy-toward-seen random for peeking 1,000 classes is used.} \label{tb_real_predicted_PZSL}
\small
\begin{tabular}{c|c|c|c|c|c}
\text{Exemplar} &  \multicolumn{5}{|c}{Flat Hit@K} \\ \cline{2-6}
\text{type} &  1 & 2  & 5 & 10 & 20 \\ \hline
Predicted & 3.1  &  4.6  &  7.5  &  10.8  &  15.6 \\ \hline
Real & 3.1 &  4.6  &  7.5  &  10.8  &  15.6 \\ \hline
\end{tabular}
\end{table} 

What should we use as visual exemplars for peeked classes? There are two options. The first option is to use their real exemplars based on (a few) peeked instances. The second option is to use their predicted visual exemplars (where real exemplars are used to learn the predictor). As the training error of the (non-linear) regressor is quite low, we observe an unnoticeable difference in zero-shot performance between the two options. For instance, in Table~\ref{tb_real_predicted_PZSL}, the two set of numbers match when the number of peeked classes is 1,000.

\section{Additional analysis on dimension for PCA}
\label{sSupplPCA}

\begin{table*}
\centering
\caption{\small Accuracy of \mtc{EXEM (1NN)} on \textbf{AwA}, \textbf{CUB}, and \textbf{SUN} when predicted exemplars are from original visual features (No PCA) and PCA-projected features (PCA with $\cst{d}$ = 1024, 500, 200, 100, 50, 10).}
\label{tSupplPCA}
\small
\begin{tabular}{c|c|c|c|c|c|c|c}
Dataset & No PCA & PCA & PCA & PCA & PCA & PCA & PCA \\ 
name & $\cst{d}$ = 1024  & $\cst{d}$ = 1024 & $\cst{d}$ = 500 & $\cst{d}$ = 200 & $\cst{d}$ = 100 & $\cst{d}$ = 50 & $\cst{d}$ = 10\\ \hline
\textbf{AwA} & \textbf{77.8} & 76.2 & 76.2 & 76.0 & 75.8 & 76.5 & 73.4\\ \hline
\textbf{CUB} & 55.1 & 56.3 & 56.3& \textbf{58.2} & 54.7 & 54.1 & 38.4\\ \hline
\textbf{SUN} & 69.2 & \textbf{69.6} & \textbf{69.6}& \textbf{69.6} & 69.3 & 68.3 & 55.3\\ \hline
\end{tabular}
\end{table*}
   
To better understand a trade-off between running time and ZSL performance, we expand Table~7 of the main text with more values for projected PCA dimensions $\cst{d}$. In Table~\ref{tSupplPCA}, we see that our approach is extremely robust. With $\cst{d}$=50, it still outperforms all baselines (cf. Table 3) on \textbf{AwA} and \textbf{SUN}; with $\cst{d}$=100, on all 3 datasets. Moreover, our method works reasonably over a wide range of (large enough) $\cst{d}$ on all datasets.

\section{Details on multi-layer perceptron}
\label{sSupplMLP}

We follow the notations defined at the beginning of Sect.~2 and Sect.~2.1 of the main text.
Similar to \cite{ZhangXG16}, our multi-layer perceptron is of the form:
\begin{equation} \label{eqMLP}
\frac{1}{\cS} \sum_{c=1}^{\cS} \twonorm{\vv_c - \mW_2 \cdot \textrm{ReLU}(\mW_1 \cdot \va_c)} + \lambda \cdot R(\mW_1, \mW_2),
\end{equation}
where $R$ denotes the $\ell_2$ regularization, $\cS$ is the number of seen classes, $\vv_c$ is the visual exemplar of class $c$, $\va_c$ is the semantic representation of class $c$, and the weights $\mW_1$ and $\mW_2$ are parameters to be optimized. 

Following \cite{ZhangXG16}, we randomly initialize the weights $\mW_1$ and $\mW_2$, and set the number of hidden units for \textbf{AwA} and \textbf{CUB} to be 300 and 700, respectively. We use Adam optimizer with a learning rate 0.0001 and minibatch size of $\cS$. We tune $\lambda$ on the same splits of data as in other experiments with class-wise CV (Sect.~\ref{sSupplHPTuning}). Our code is implemented in TensorFlow \cite{tensorflow}.

\end{document}